\pdfoutput=1

\documentclass[11pt]{article}

\usepackage{acl}

\usepackage{times}
\usepackage{latexsym}
\usepackage{booktabs}
\usepackage{amsmath}
\usepackage{multirow}

\usepackage[T1]{fontenc}

\usepackage[utf8]{inputenc}
\usepackage{graphicx}

\usepackage{microtype}

\newcommand{\tablewhitespace}{\addlinespace[0.3em]}

\newcommand{\OPT}[0]{{OPT-175B}}
\newcommand{\davinci}[0]{{Davinci}}

\newcommand{\biggpt}[1]{{GPT-3#1}}

\title{OPT: Open Pre-trained Transformer Language Models}

\author{
{\bf Susan Zhang}\thanks{\ \ Equal contribution.
}\footnotemark[1]{\normalfont ,} 
{\bf Stephen Roller}\footnotemark[1]{\normalfont ,} 
{\bf Naman Goyal}\footnotemark[1]{\normalfont ,} 
\\
{\bf Mikel Artetxe}{\normalfont ,}
{\bf Moya Chen}{\normalfont ,}
{\bf Shuohui Chen}{\normalfont ,}
{\bf Christopher Dewan}{\normalfont ,}
{\bf Mona Diab}{\normalfont ,}
{\bf Xian Li}{\normalfont ,} \\
{\bf Xi Victoria Lin}{\normalfont ,}
{\bf Todor Mihaylov}{\normalfont ,}
{\bf Myle Ott}\thanks{\ \ Work done while at Meta AI.}\footnotemark[2]{\normalfont ,}
{\bf Sam Shleifer}\footnotemark[2]{\normalfont ,} 
{\bf Kurt Shuster}{\normalfont ,} 
{\bf Daniel Simig}{\normalfont ,} \\
{\bf Punit Singh Koura}{\normalfont ,}
{\bf Anjali Sridhar}{\normalfont ,}
{\bf Tianlu Wang}{\normalfont ,}
{\bf Luke Zettlemoyer}{\normalfont } \\
\tablewhitespace
Meta AI\\
\tablewhitespace
\texttt{\{susanz,roller,naman\}@fb.com}
}
\begin{document}
\maketitle
\begin{abstract}
Large language models, which are often trained for hundreds of thousands of compute days, have shown remarkable capabilities for zero- and few-shot learning. Given their computational cost, these models are difficult to replicate without significant capital.  For the few that are available through APIs, no access is granted to the full model weights, making them difficult to study. We present Open Pre-trained Transformers (OPT), a suite of decoder-only pre-trained transformers ranging from 125M to 175B parameters, which we aim to fully and responsibly share with interested researchers. We show that \OPT{} is comparable to \biggpt{},\footnote{Following \citet{brown2020gpt3}, we use \biggpt{} to  refer to both the 175B model and the smaller scale models as well.} while requiring only 1/7th the carbon footprint to develop.  We are also releasing our logbook detailing the infrastructure challenges we faced, along with code for experimenting with all of the released models.
\end{abstract}

\section{Introduction}

Large language models (LLMs) trained on massive text collections have shown surprising emergent capabilities to generate text and perform zero- and few-shot learning~\cite{brown2020gpt3,J1WhitePaper,megatron2022,gopher2022,palm2022}. While in some cases the public can interact with these models through paid APIs, full model access is currently limited to only a few highly resourced labs.\footnote{Exceptions include work by EleutherAI, who released dense models up to 20B in size~\cite{eleutherai2022}, Salesforce \cite{Nijkamp2022ACP}, and Meta AI, who released dense models up to 13B and sparse models up to 1.1T~\cite{1TMoE2021}. There is also ongoing work from the BigScience workshop (\url{https://bigscience.huggingface.co/}), which aims to open source very large multilingual language models and datasets.} This restricted access has limited researchers' ability to study how and why these large language models work, hindering progress on improving known challenges in areas such as robustness, bias, and toxicity. 

In this technical report, we present Open Pre-trained Transformers (OPT), a suite of decoder-only pre-trained transformers ranging from 125M to 175B parameters, which we aim to fully and responsibly share with interested researchers. We train the OPT models to roughly match the performance and sizes of the GPT-3 class of models, while also applying the latest best practices in data collection and efficient training. Our aim in developing this suite of OPT models is to enable reproducible and responsible research at scale, and to bring more voices to the table in studying the impact of these LLMs.  Definitions of risk, harm, bias, and toxicity, etc., should be articulated by the collective research community as a whole, which is only possible when models are available for study.  

We are releasing all of our models between 125M and 66B parameters, and will provide full research access to OPT-175B upon request. Access will be granted to academic researchers; those affiliated with organizations in government, civil society, and academia; and those in industry research laboratories. 
We are also releasing both the logbook of our model creation as well as our codebase, {metaseq},\footnote{\url{https://github.com/facebookresearch/metaseq}} which enabled training \OPT{} on 992 80GB A100 GPUs, reaching ~147 TFLOP/s utilization per GPU.  From this implementation, and from using the latest generation of NVIDIA hardware, we are able to develop \OPT{} using only 1/7th the carbon footprint of \biggpt{.}  While this is a significant achievement, the energy cost of creating such a model is still nontrivial, and repeated efforts to replicate a model of this size will only amplify the growing compute footprint of these LLMs.

We believe the entire AI community — academic researchers, civil society, policymakers, and industry — must work together to develop clear guidelines around responsible AI in general and responsible LLMs in particular, given their centrality in many downstream language applications. A much broader segment of the AI community needs access to these models in order to conduct reproducible research and collectively drive the field forward. With the release of OPT-175B and smaller-scale baselines, we hope to increase the diversity of voices defining the ethical considerations of such technologies.

\section{Method}
\label{sec:method}

\subsection{Models}
\label{sec:models}

We present results on eight Transformer language models ranging from 125 million to 175 billion parameters. Architectural details are displayed in Table \ref{tab:model_sizes}. In the interest of transparency, and to reduce risk of training instabilities, our models and hyperparameters largely follow \citet{brown2020gpt3}, with variations in batch size mostly to obtain increased computational efficiency.

\begin{table}[t]
    \centering
    \begin{tabular}{lrrrrr}
    \toprule
    {\bf Model} & {\bf \#L} & {\bf \#H}  & {\bf d$_{\text{model}}$} & {\bf LR} & {\bf Batch} \\
    \midrule
        125M    & 12 & 12 & 768   & $6.0e{-4}$ & 0.5M \\
        350M    & 24 & 16 & 1024  & $3.0e{-4}$ & 0.5M \\
        1.3B    & 24 & 32 & 2048  & $2.0e{-4}$ & 1M \\
        2.7B    & 32 & 32 & 2560  & $1.6e{-4}$ & 1M \\
        6.7B    & 32 & 32 & 4096  & $1.2e{-4}$ & 2M \\
        13B     & 40 & 40 & 5120  & $1.0e{-4}$ & 4M \\
        30B     & 48 & 56 & 7168  & $1.0e{-4}$ & 4M \\
        66B     & 64 & 72 & 9216  & $0.8e{-4}$ & 2M \\
        175B    & 96 & 96 & 12288 & $1.2e{-4}$ & 2M\\
    \bottomrule
    \end{tabular}
    \caption{{\bf Model architecture details.} We report the number of layers (\#L),
    number of attention heads (\#H), and the embedding size ({d$_{\text{model}}$}).
    We also report the peak Learning Rate (LR) and global batch size in number of tokens (Batch).
    }
    \label{tab:model_sizes}
\end{table}
\subsection{Training Setup}
\label{sec:training_setup}
For weight initialization, we follow the same settings provided in the Megatron-LM codebase,\footnote{\url{https://github.com/NVIDIA/Megatron-LM/blob/main/examples/pretrain_gpt3_175B.sh}} using a normal distribution with zero mean and standard deviation of 0.006.  Standard deviation for output layers are scaled by a $1.0/\sqrt{2 L}$ term where $L$ is the total number of layers. All bias terms are initialized as 0, and all models are trained with ReLU activation and a sequence length of 2048.

We use an AdamW optimizer \cite{AdamW} with $(\beta_1,\beta_2)$ set to $(0.9, 0.95)$, and weight decay of $0.1$.  We follow a linear learning rate schedule, warming up from $0$ to the maximum learning rate over the first 2000 steps in OPT-175B, or over 375M tokens in our smaller baselines, and decaying down to 10\% of the maximum LR over 300B tokens.  A number of mid-flight changes to LR were also required (see Section~\ref{sec:training_process}).  Our batch sizes range from 0.5M to 4M depending on the model size (see Table \ref{tab:model_sizes}) and is kept constant throughout the course of training.

We use a dropout of 0.1 throughout, but we do not apply any dropout to embeddings. We clip gradient norms at 1.0, except for some mid-flight changes that reduce this threshold down from 1.0 to 0.3 (see Section~\ref{sec:training_process}).  We also include a gradient predivide factor to reduce the risk of over/underflows when computing the gradient across all ranks (splitting the division by the world size of $N$ into two division operations by $\sqrt{N}$).

\subsection{Pre-training Corpus}
\label{sec:training_dataset}
The pre-training corpus contains a concatenation of datasets used in RoBERTa \cite{liu2019roberta}, the Pile \cite{thepile}, and PushShift.io Reddit \cite{reddit2020, roller-etal-2021-recipes}. All corpora were previously collected or filtered to contain predominantly English text, but a small amount of non-English data is still present within the corpus via CommonCrawl.

We removed duplicated documents across all datasets by filtering out documents via MinhashLSH \cite{rajaraman2011mining} with a Jaccard similarity $\geq.95$. We found the Pile was particularly full of duplicate documents, and advise future researchers using the Pile to perform additional de-duplication processing.

We tokenize all corpora using the GPT-2 byte level BPE tokenizer \cite{sennrich2016bpe,radford2019language,brown2020gpt3}. Our final corpus contains roughly 180B tokens.

\paragraph{RoBERTa} We included the BookCorpus \cite{books2015} and Stories \cite{ccstories2018} subsets of the RoBERTa corpus and utilized an updated version of CCNews, containing news stories crawled through September 28, 2021. This CCNews v2 corpus was preprocessed the same way as the original RoBERTa CCNews \cite{liu2019roberta}.

\paragraph{The Pile} We included a subset of the Pile \cite{thepile}, including: CommonCrawl, DM Mathematics,  Project Gutenberg, HackerNews, OpenSubtitles, OpenWebText2, USPTO and Wikipedia. Other subsets of the Pile were eliminated as we found they increased the risk of instabilities, as measured by tendency to cause spikes in gradient norms at the 1.3B scale, or were otherwise deemed unsuitable.  All subsets went through additional ad-hoc whitespace normalization.

\paragraph{PushShift.io Reddit} We included a subset of the Pushshift.io corpus produced by \citet{reddit2020} and previously used by \citet{roller-etal-2021-recipes}. To convert the conversational trees into language-model-accessible documents, we extracted the longest chain of comments in each thread and discarded all other paths in the tree. This reduced the corpus by about 66\%.

\subsection{Training Efficiency}
\label{sec:training_efficiency}
We trained \OPT{} on 992 80GB A100 GPUs, by utilizing Fully Sharded Data Parallel \cite{1TMoE2021} with Megatron-LM Tensor Parallelism \cite{shoeybi2019megatron}. We achieve utilization of up to 147 TFLOP/s per GPU. We keep Adam state in FP32, since we shard it across all hosts, while the model weights remained in FP16. To avoid underflows, we used dynamic loss scaling, as described in \citet{micikevicius2017mixed}.

\subsection{Training Processes}
Here we describe significant training process adjustments that arose during \OPT{} pre-training.

\label{sec:training_process}
\paragraph{Hardware Failures} 
We faced a significant number of hardware failures in our compute cluster while training \OPT{}. In total, hardware failures contributed to at least 35 manual restarts and the cycling of over 100 hosts over the course of 2 months. During manual restarts, the training run was paused, and a series of diagnostics tests were conducted to detect problematic nodes. Flagged nodes were then cordoned off and training was resumed from the last saved checkpoint. Given the difference between the number of hosts cycled out and the number of manual restarts, we estimate 70+ automatic restarts due to hardware failures.

\begin{figure}[t]
    \centering
    \includegraphics[width=0.98\linewidth]{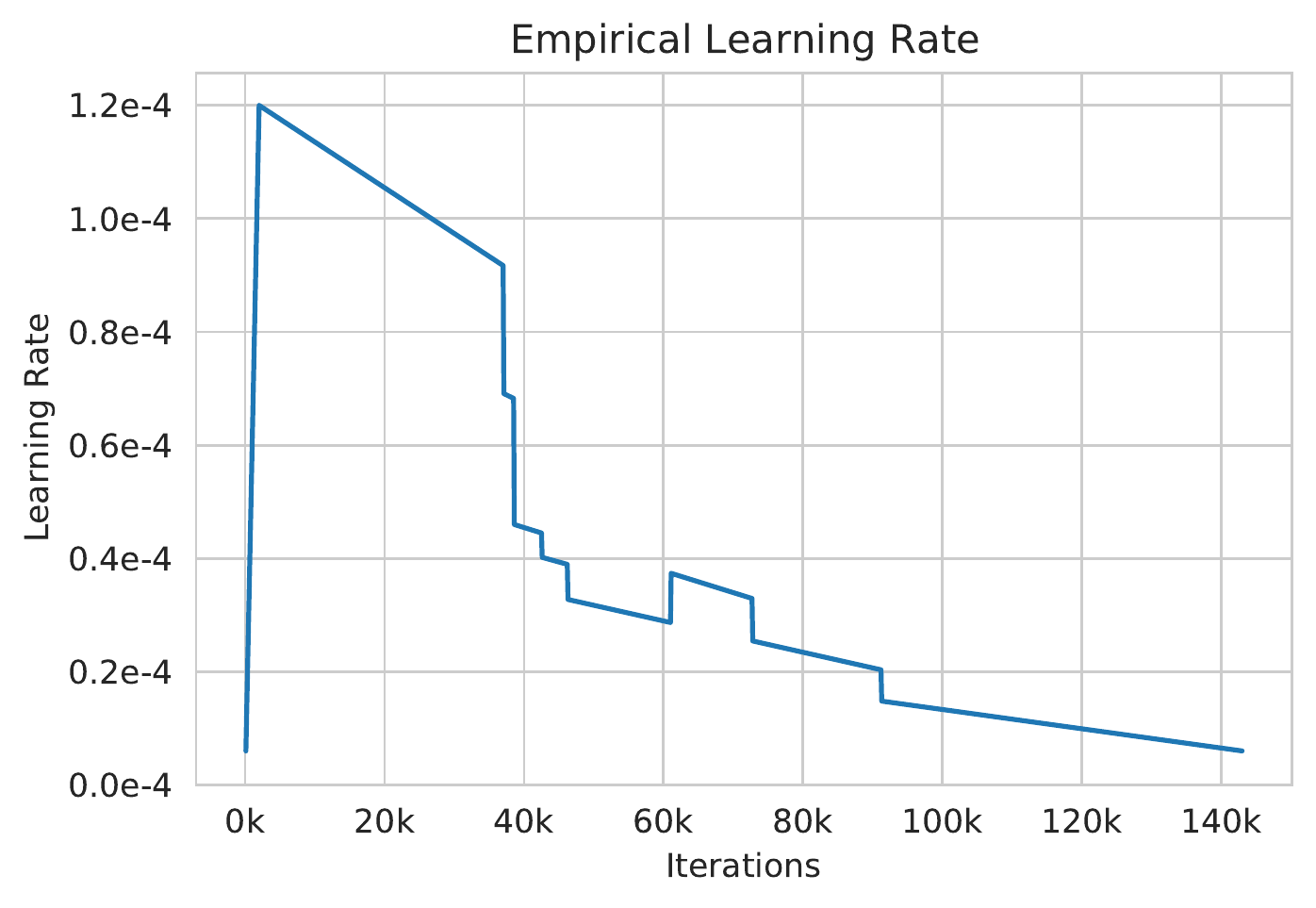}
    \caption{\textbf{Empirical LR schedule.} We found that lowering learning rate was helpful for avoiding instabilities.}
    \label{fig:lr}
\end{figure}

\begin{figure}
    \centering
    \includegraphics[width=0.98\linewidth]{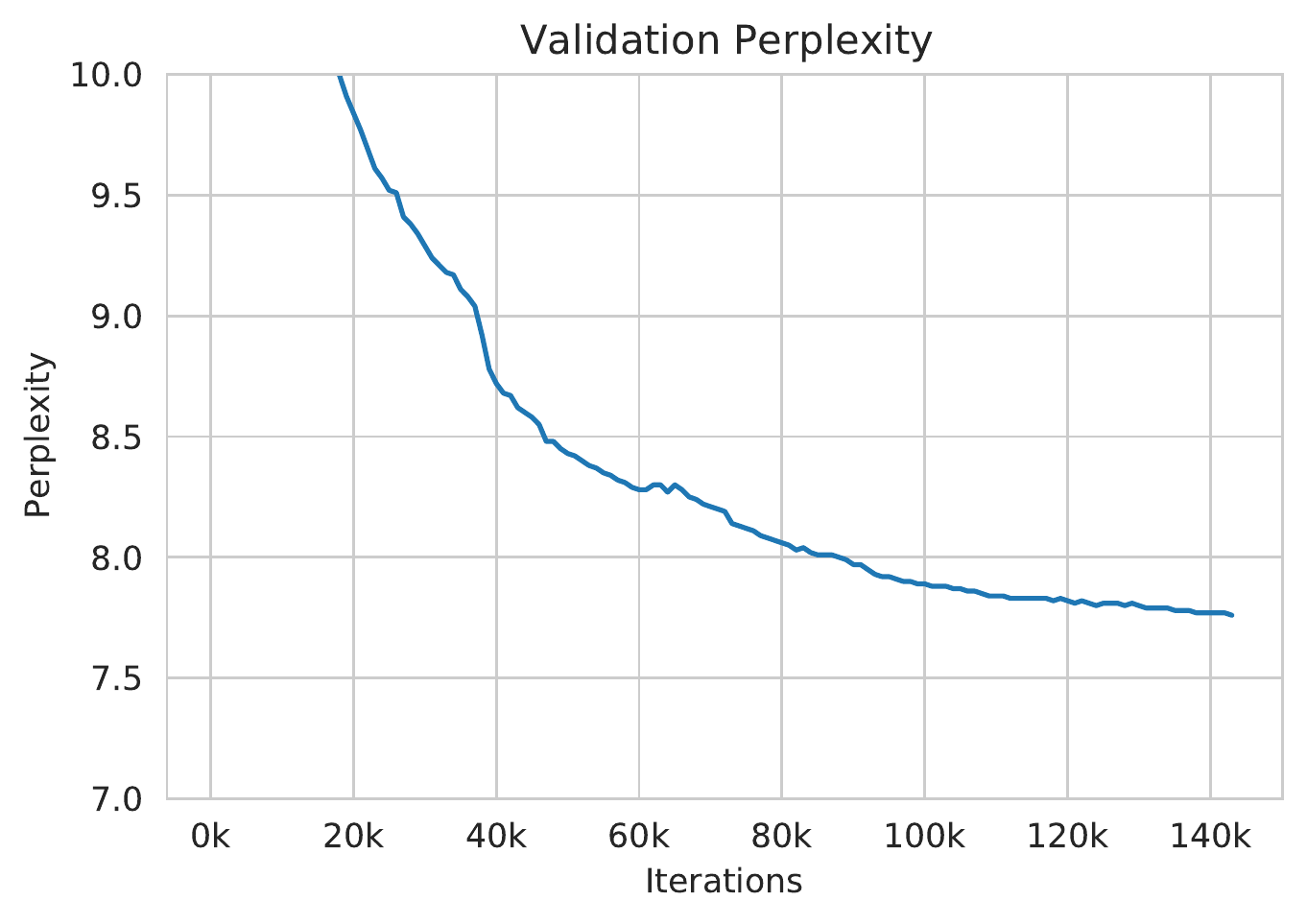}
    \caption{\textbf{Validation Perplexity.} Our mid-flight LR
    changes had clear effects on validation perplexity.}
    \label{fig:validation_ppl}
\end{figure}

\paragraph{Loss Divergences}
Loss divergences were also an issue in our training run. When the loss diverged, we found that lowering the learning rate and restarting from an earlier checkpoint allowed for the job to recover and continue training.  We noticed a correlation between loss divergence, our dynamic loss scalar crashing to 0, and the $l^2$-norm of the activations of the final layer spiking. These observations led us to pick restart points for which our dynamic loss scalar was still in a ``healthy'' state ($\geq1.0$), and after which our activation norms would trend downward instead of growing unboundedly.  Our empirical LR schedule is shown in Figure~\ref{fig:lr}. Early in training, we also noticed that lowering gradient clipping from 1.0 to 0.3 helped with stability; see our released logbook for exact details.
Figure~\ref{fig:validation_ppl} shows our validation loss with respect to training iterations.

\paragraph{Other Mid-flight Changes} 
We conducted a number of other experimental mid-flight changes to handle loss divergences. These included: switching to vanilla SGD (optimization plateaued quickly, and we reverted back to AdamW); resetting the dynamic loss scalar (this helped recover some but not all divergences); and switching to a newer version of Megatron (this reduced pressure on activation norms and improved throughput).

\section{Evaluations}
\label{sec:main_results}

\subsection{Prompting \& Few-Shot}
\label{sec:prompt_and_fewshot}

We evaluate our model on 16 standard NLP tasks utilized in the literature: HellaSwag \cite{hellaswag}, StoryCloze \cite{storycloze}, PIQA \cite{bisk2020piqa}, ARC Easy and Challenge \cite{arc_challenge}, OpenBookQA \cite{openbookqa}, WinoGrad \cite{levesque2011winograd}, WinoGrande \cite{winogrande}, and SuperGLUE \cite{wang2019superglue}. We follow GPT-3 \cite{brown2020gpt3} by using their prompts and overall experimental setup. We compare primarily to GPT-3, having aimed to re-implement their  evaluation settings, but include reported performance of other LLMs on a per-task basis when available \cite{J1WhitePaper,gopher2022,chinchilla2022,eleutherai2022}

We report performance in accuracy (omitting F1 for MultiRC and ReCoRD for consistency in evaluation metrics). For the Winograd Schema Challenge (WSC) task in the SuperGLUE benchmark, we follow \cite{brown2020gpt3} and formulate the task as multiple choice questions, which is known to affect performance \cite{liu2020precise}.

\begin{figure}
    \centering
    \includegraphics[width=0.98\linewidth]{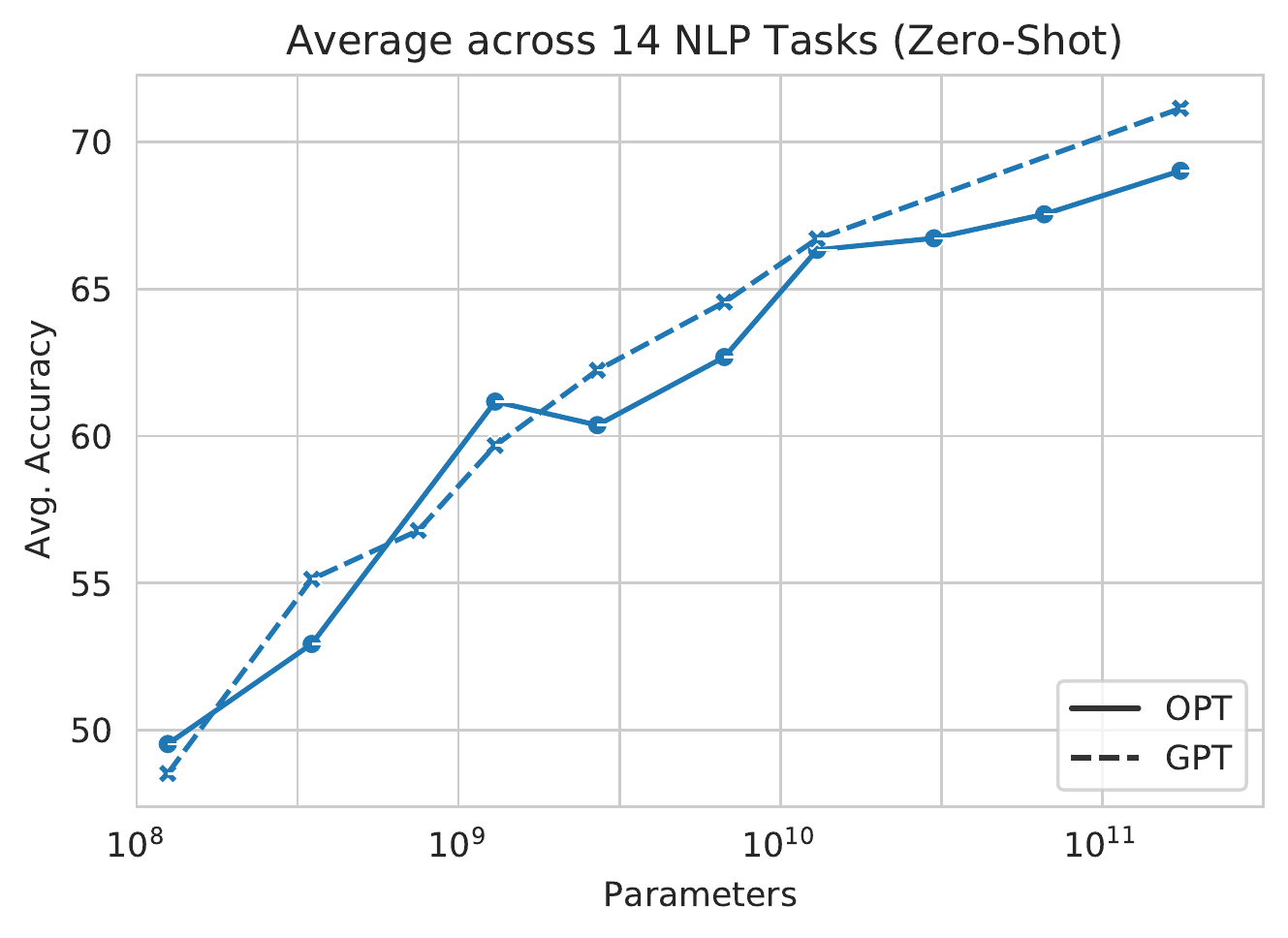}
    \caption{{\bf Zero-shot NLP Evaluation Averages}. Across a variety of tasks and model sizes, OPT largely matches the reported averages of GPT-3. However, performance varies greatly per task: see Appendix~\ref{app:full_evals}.}
    \label{fig:zeroshot}
\end{figure}

\paragraph{Zero-shot} Overall average zero-shot performance across all 14 tasks may be seen in Figure~\ref{fig:zeroshot}. Overall, we see our average performance follows the trend of GPT-3. However, performance can vary radically across the tasks: for a full breakdown, see Appendix~\ref{app:full_evals}.  Note that we intentionally removed MultiRC and WIC from these averages, as these datasets seem to systematically favor GPT-3 or OPT disproportionately.

Our performance roughly matched GPT-3 for 10 tasks, and underperformed in 3 tasks (ARC Challenge and MultiRC). In 3 tasks (CB, BoolQ, WSC), we find both GPT and OPT models display unpredictable behavior with respect to scale, likely due to the small size of the validation set in these 3 tasks (56, 277, and 104 examples, respectively). In WIC, we see that the OPT models always outperform the GPT-3 models, though the numbers reported by \citet{brown2020gpt3} also seem questionable, given WIC being a binary classification task.\footnote{\citet{brown2020gpt3} reports 0\% accuracy on WIC, which implies 100\% accuracy if the classification was inverted.} For MultiRC, we are unable to replicate the GPT-3 results using the \davinci{} API\footnote{\url{https://beta.openai.com/docs/engines/overview}} within our evaluation setup, suggesting differences in the methods of evaluation on this task. For BoolQ and WSC, we note that both OPT and GPT models seem to hover around majority-class accuracy, suggesting small perturbations in probability masses may be dominating the evaluations.

Chinchilla \cite{chinchilla2022} and Gopher \cite{gopher2022} perform roughly consistently with others for their parameter sizes, while PaLM \cite{palm2022} generally performs better across all settings, even when controlling for number of parameters. We speculate the high performance of PaLM comes predominantly from higher quality and diversity of pre-training data.

\begin{figure}[t]
    \centering
    \includegraphics[width=0.98\linewidth]{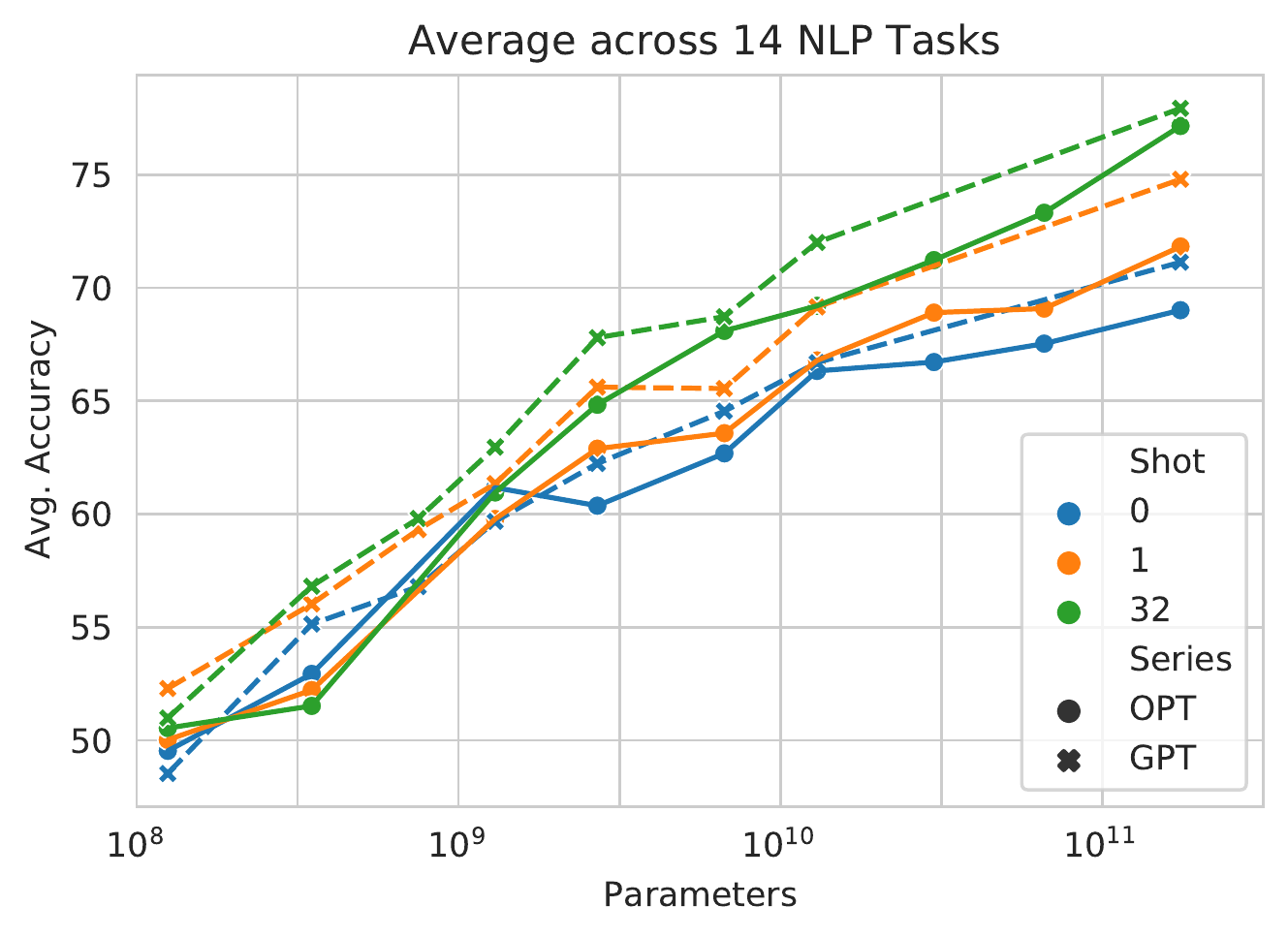}
    \caption{\textbf{Multi-shot performance}. OPT performance for one- and few-shot lags behind GPT-3 models, but performance depends heavily per task; see Appendix~\ref{app:full_evals}.}
    \label{fig:multishot}
\end{figure}

\paragraph{One-shot and Few-shot} Average multi-shot in-context performance is shown in Figure~\ref{fig:multishot} (again, omitting MultiRC and WIC), with detailed performances shown in Appendix~\ref{app:full_evals}. Across the average of all metrics, we find that OPT models perform similarly to GPT-3 models. However, as with zero-shot, breaking down these results per task shows a different story: in the same set of 10 datasets as zero-shot, we see similar performance across the two models. Some of the remaining datasets show inconsistent performance with respect to model size for both OPT and GPT-3 models (BoolQ, CB, WSC, RTE). In MultiRC, we consistently see underperformance of OPT models compared to GPT-3 models. Similar to our zero-shot evaluation, we hypothesize our one- and few-shot evaluation setup may differ significantly from \citet{brown2020gpt3}.

\subsection{Dialogue}
\label{sec:dialogue}

Given that LLMs are known to be an integral component of modern dialogue models \cite{adiwardana2020meena,roller-etal-2021-recipes,thoppilan2022lamda,gopher2022,palm2022}, we additionally evaluate \OPT{} on several open source dialogue datasets. In particular, we follow \citet{roller-etal-2021-recipes}, and evaluate on ConvAI2 \cite{dinan2019second}, Wizard of Wikipedia \cite{dinan2018wizard}, Empathetic Dialogues \cite{rashkin2019empathetic}, and Blended Skill Talk \cite{smith2020bst}. We additionally evaluate on the more recent Wizard of Internet dataset \cite{komeili2021woi}.
We focus our comparisons primarily against existing open source dialogue models including the fine-tuned BlenderBot~1 \cite{roller-etal-2021-recipes} and its pre-training counterpart Reddit 2.7B. We also compare against the fine-tuned R2C2 BlenderBot, a 2.7B parameter BlenderBot-like model trained by \citet{shuster2022language}.

We report Perplexity and Unigram F1 (UF1) overlap, following the metrics of the ConvAI2 competition \cite{dinan2019second}. To control for different tokenization in each of the models, we normalize all perplexities to be in the space of the GPT-2 tokenizer \cite{radford2019language}. We also note which models are supervised with respect to these dialogue tasks and which are unsupervised. For \OPT{}, all generations are performed using greedy decoding up to a maximum of 32 tokens. We do not attempt to prompt the model at all except for alternating ``Person 1:'' and ``Person 2:'' lines of dialogue. The remaining models use the generation parameters found in BlenderBot~1.

\begin{table*}[t]
    \centering
    \begin{tabular}{llrrrrrrrrrr}
    \toprule
    & & \multicolumn{5}{c}{Perplexity ($\downarrow$)} & \multicolumn{5}{c}{Unigram F1 ($\uparrow$)}\\
    \cmidrule(lr){3-7} \cmidrule(lr){8-12}
    {\bf Model} & {\bf Eval} & {\bf C2} & {\bf WW} & {\bf ED} & {\bf BST} & {\bf WoI} & {\bf C2} & {\bf WW} & {\bf ED} & {\bf BST} & {\bf WoI}\\
    \midrule
         Reddit 2.7B             & Unsup. &  18.9  &    21.0 & 11.6   & 17.4    & 18.0   & .126 & .133 & .135 & .133 & .124 \\
         BlenderBot 1            & Sup.   &{\bf10.2}  &    12.5 &{\bf9.0}& 11.9    & 14.7   & .183 & .189 & .192 & .178 & .154\\
         R2C2 BlenderBot             & Sup.   &  10.5  &   {\bf 12.4} &  9.1   &{\bf11.7}& 14.6   &{\bf.205}&{\bf.198}&{\bf.197}&{\bf.186}&{\bf.160}\\
         \tablewhitespace
         \OPT                    & Unsup. &  10.8 & 13.3 & 10.3  &  12.1 & {\bf 12.0} & .185 & .152 & .149 & .162 & .147\\
    \bottomrule
    \end{tabular}
    \caption{{\bf Dialogue Evaluations.} OPT-175B, in a fully unsupervised setting, performs competitively against fully supervised models.}
    \label{tab:dialogue_evals}
\end{table*}

Results are shown in Table~\ref{tab:dialogue_evals}. We see that \OPT{} significantly outperforms the also-unsupervised Reddit 2.7B model on all tasks, and performs competitively with the fully supervised BlenderBot~1 model, especially in the ConvAI2 dataset. On the Wizard-of-Internet dataset, which is fully unsupervised for all models, we see that \OPT{} obtains the lowest perplexity but still has lower UF1 than the models with Wizard-of-Wikipedia supervision.

We were somewhat surprised that the evaluations of the unsupervised \OPT{} model were as competitive as BlenderBot~1 on the ConvAI2 dataset. This may indicate leakage of the ConvAI2 dataset into the general pre-training corpus or even into the validation data as evaluated in Table~\ref{tab:dialogue_evals}. To address concerns of leakage, we searched our pre-training corpus for the first conversation in the ConvAI2 dataset, but we did not find any overlap.
We additionally evaluated \OPT{} on the ConvAI2 hidden test set, which has never been publicly released, and achieved 10.7 ppl and .185 UF1, matching the performance of the validation set. Furthermore, we evaluated \OPT{} on a subset of the ConvAI2-like MultiSessionChat (MSC) dataset \cite{xu2021beyond} and obtained a perplexity of 9.7 and UF1 of .177, indicating the model is generalizing well across multiple PersonaChat-like datasets. Since both MSC and WoI datasets were released \textit{after} the CommonCrawl snapshot used in pre-training corpus, there is minimal risk of leakage. We conclude that \OPT{} has a strong ability to maintain a consistent persona across conversations, a behavior also highlighted in LaMDA \cite{thoppilan2022lamda}.

\section{Bias \& Toxicity Evaluations}
\label{sec:rai_evals}

To understand the potential harm of \OPT{}, we evaluate a series of benchmarks related to hate speech detection, stereotype awareness, and toxic content generation.  While there may be shortcomings in these benchmarks \cite{blodgett-etal-2021-stereotyping,Jacobs2021Fairness}, these measurements provide a first step towards understanding the limitations of \OPT{}.  We compare primarily against GPT-3 \davinci{}, as these benchmarks were not yet available to be included in  \citet{brown2020gpt3}.

\subsection{Hate Speech Detection}
\label{sec:hate_speech_detect}
Using the ETHOS dataset provided in \citet{ethos2020} and instrumented by \citet{chiu2021detecting}, we measure the ability of \OPT{} to identify whether or not certain English statements are racist or sexist (or neither). In the zero-, one-, and few-shot binary cases, the model is presented with text and asked to consider whether the text is racist or sexist and provide a yes/no response. In the few-shot multiclass setting, the model is asked to provide a yes/no/neither response.

Results are presented in Table~\ref{tab:hatespeech}.
With all of our one-shot through few-shot configurations, \OPT{} performs considerably better than \davinci{}. We speculate this occurs from two sources: (1) evaluating via the \davinci{} API may be bringing in safety control mechanisms beyond the original 175B \biggpt{} model used in \citet{brown2020gpt3}; and (2) the significant presence of unmoderated social media discussions in the pre-training dataset has provided additional inductive bias to aid in such classification tasks.

\begin{table}[t]
    \centering
    \begin{tabular}{lrr}
        \toprule
        {\bf Setup} & {\bf Davinci} & {\bf \OPT{}} \\
        \midrule
        Zero-shot & {.628} & {\bf .667}\\
        One-shot  & {.616} & {\bf .713}\\
        Few-shot (binary) & {.354} & {\bf .759}\\
        Few-shot (multiclass) & {.672} & {\bf.812}\\
        \bottomrule
    \end{tabular}
    \caption{{\bf Hate speech detection.} F1 scores of detecting hate speech between \davinci{} and \OPT{}. \OPT{} considerably outperforms \davinci{} in all settings.
    }
    \label{tab:hatespeech}
\end{table}

\subsection{CrowS-Pairs}
Developed for masked language models, CrowS-Pairs \cite{nangia2020crows} is a crowdsourced benchmark aiming to measure intrasentence level biases in 9 categories: gender, religion, race/color, sexual orientation, age, nationality, disability, physical appearance, and socioeconomic status. Each example consists of a pair of sentences representing a stereotype, or anti-stereotype, regarding a certain group, with the goal of measuring model preference towards stereotypical expressions.  Higher scores indicate higher bias exhibited by a model.

When compared with \davinci{} in Table~\ref{tab:crowspairs}, \OPT{} appears to exhibit more stereotypical biases in almost all categories except for religion.  Again, this is likely due to differences in training data;  \citet{nangia2020crows} showed that Pushshift.io Reddit corpus has a higher incidence rate for stereotypes and discriminatory text than other corpora (e.g.~Wikipedia). Given this is a primary data source for \OPT{}, the model may have learned more discriminatory associations, which directly impacts its performance on CrowS-Pairs. 

\begin{table}[t]
    \centering
    \begin{tabular}{lrr}
\toprule
Category &  {\bf GPT-3} & {\bf \OPT{}}\\
\midrule
Gender    & {\bf 62.6} & 65.7\\
Religion & 73.3	& {\bf 68.6}\\
Race/Color & {\bf 64.7} & 68.6\\
Sexual orientation & {\bf 76.2} & 78.6\\
Age & {\bf 64.4} & 67.8\\
Nationality & {\bf 61.6} & 62.9\\
Disability & {\bf 76.7} & {\bf 76.7}\\
Physical appearance & {\bf 74.6} & {76.2}\\
Socioeconomic status & {\bf 73.8} & 76.2\\
\cmidrule(lr){1-3}
Overall & {\bf 67.2} & 69.5\\
\bottomrule
    \end{tabular}
    \caption{{\bf CrowS-Pairs evaluation.} Lower is better for all categories, indicating more fairness. The \OPT{} model
    performs worse than \davinci{} in most categories.}
    \label{tab:crowspairs}
\end{table}

\subsection{StereoSet}

Following \citet{J1WhitePaper} and \citet{1TMoE2021}, we use StereoSet \cite{nadeem2020stereoset} to measure stereotypical bias across 4 categories: profession, gender, religion, and race.  In addition to intrasentence measurement (similar to CrowS-Pairs), StereoSet includes  measurement at the intersentence level to test a model's ability to incorporate additional context.  To account for a potential trade-off between bias detection and language modeling capability, StereoSet includes two metrics: Language Modeling Score (LMS) and Stereotype Score (SS), which are then combined to form the Idealized Context Association Test score (ICAT). Unlike \citet{J1WhitePaper}, we normalize scores by token count, rather than character count, which they report improves metrics for several models.

Results are shown in Table~\ref{tab:stereoset}. We see that \davinci{} and \OPT{} exhibit similar scores on aggregate (overall ICAT is very close between the two). In particular, \davinci{} outperforms in the areas of profession and race, while \OPT{} outperforms in the areas of Gender and Religion. \OPT{} performs better across the board on the SS metric, while \davinci{} generally outperforms on the LMS metric.

\begin{table}[t]
    \centering
    \begin{tabular}{lrrr}
        \toprule
        \multicolumn{2}{l}{{\bf Category}} & {\bf Davinci} & {\bf \OPT{}} \\
        \midrule
        \multirow{3}{*}{Prof.}   & LMS ($\uparrow$)   & {\bf 78.4} & 74.1\\
                                 & SS  ($\downarrow)$ & {    63.4} & {\bf 62.6}\\
                                 & ICAT ($\uparrow$)  & {\bf 57.5} & 55.4\\

        \cmidrule(lr){1-4}             
        \multirow{3}{*}{Gend.}   & LMS ($\uparrow$)   & {\bf 75.6} & 74.0\\
                                 & SS  ($\downarrow)$ & {    66.5} & {\bf 63.6}\\
                                 & ICAT ($\uparrow$)  & {    50.6} & {\bf 53.8}\\

        \cmidrule(lr){1-4}             
        \multirow{3}{*}{Reli.}   & LMS ($\uparrow$)   & {    80.8} & {\bf 84.0}\\
                                 & SS  ($\downarrow)$ & {\bf 59.0} & {\bf 59.0}\\
                                 & ICAT ($\uparrow$)  & {    66.3} & {\bf 68.9}\\

        \cmidrule(lr){1-4}             
        \multirow{3}{*}{Race}    & LMS ($\uparrow$)   & {\bf 77.0} & 74.9\\
                                 & SS  ($\downarrow)$ & {    57.4} & {\bf 56.8}\\
                                 & ICAT ($\uparrow$)  & {\bf 65.7} & 64.8\\

        \cmidrule(lr){1-4}             
        \multirow{3}{*}{Overall} & LMS ($\uparrow$)   & {\bf 77.6} & 74.8\\
                                 & SS  ($\downarrow)$ & {    60.8} & {\bf 59.9}\\
                                 & ICAT ($\uparrow$)  & {\bf 60.8} & 60.0\\

        \tablewhitespace
        \bottomrule
    \end{tabular}
    \caption{{\bf StereoSet Evaluations}. \davinci{} and \OPT{} perform similarly across all evaluations.}
    \label{tab:stereoset}
\end{table}

\subsection{RealToxicityPrompts}

We evaluate the tendency of \OPT{} to respond with toxic language via the RealToxicityPrompts \cite{gehman2020realtoxicityprompts} dataset. Following PaLM \cite{palm2022}, we sample 25 generations of 20 tokens using nucleus sampling \cite{holtzman2020nucleus} ($p=0.9$) for each of $10,000$ randomly sampled prompts from RTP, and report mean toxicity probabilities of the continuations, stratified across bucketed toxicities of the original prompts. For comparison, we report bucketed toxicity rates from \davinci{} and PaLM.

\begin{figure}[t]
    \centering
    \includegraphics[width=\linewidth]{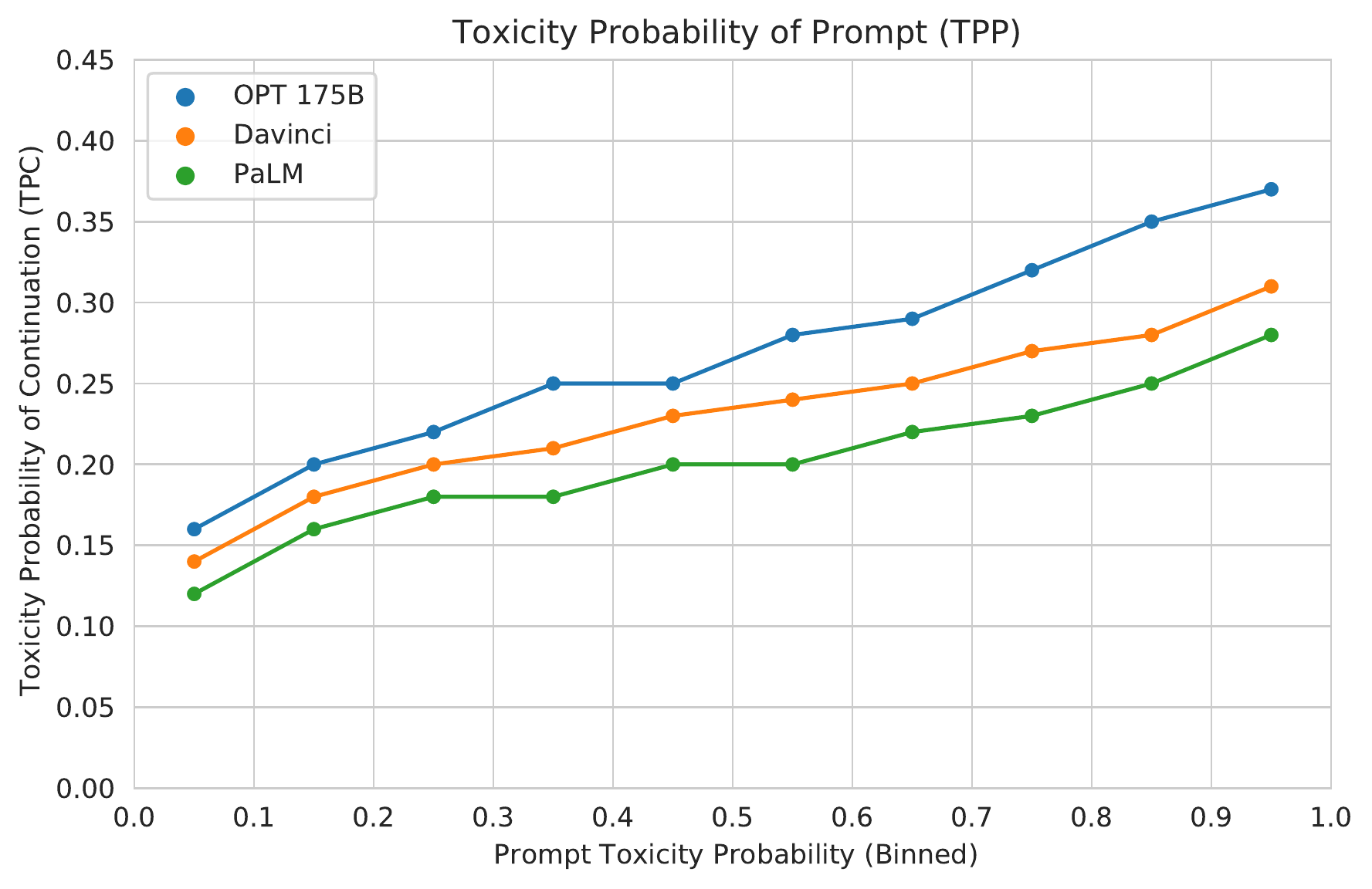}
    \caption{{\bf RealToxicityPompts}. \OPT{} is more likely to generate toxic responses than either \davinci{} or PaLM. Consistent with prior work, toxicity rates increase as prompt toxicity increases.}
    \label{fig:rtp}
\end{figure}

Results are shown in Figure~\ref{fig:rtp}. Overall, we see that \OPT{} has a higher toxicity rate than either PaLM or \davinci{}. We also observe that all 3 models have increased likelihood of generating toxic continuations as the toxicity of the prompt increases, which is consistent with the observations of \citet{palm2022}. As with our experiments in hate speech detection, we suspect the inclusion of unmoderated social media texts in the pre-training corpus raises model familiarity with, and therefore propensity to generate and detect, toxic text. This strong awareness of toxic language may or may not be desirable depending on the specific requirements of downstream applications. Future applications of \OPT{} should consider this aspect of the model, and take additional mitigations, or avoid usage entirely as appropriate.

\subsection{Dialogue Safety Evaluations}

Finally, we compare \OPT{} on two Dialogue Safety evaluations. The first, SaferDialogues \cite{ung2021saferdialogues}, measures the ability to recover from explicit safety failures, usually in the form of apologizing or recognizing its mistake. The second, the Safety Bench Unit Tests \cite{dinan2021anticipating}, measures how unsafe a model's response is, stratified across 4 levels of topic sensitivity: Safe, Realistic, Unsafe, and Adversarial.
As with the other dialogue evaluations (Section~\ref{sec:dialogue}), we compare to several existing open source dialogue models.

\begin{table}[t]
    \centering
\resizebox{\linewidth}{!}{
\begin{tabular}{lrrrrrr}
    \toprule
    & \multicolumn{2}{c}{Safe. Dia.} & \multicolumn{4}{c}{Unit Tests ($\downarrow$)}\\
    \cmidrule(lr){2-3} \cmidrule(lr){4-7}
      {\bf Model} & {\bf PPL} & {\bf F1} & {\bf Sa} & {\bf Re} & {\bf Un} & {\bf Ad}\\
     \midrule
    Reddit 2.7B & 16.2 & .140 & .300 & .261 & .450 & .439\\
    BlenderBot 1 & {\bf 12.4} & {\bf .161} & .028 & .150 & {\bf .250} & {\bf .194}\\
    R2C2 BlenderBot & 13.8 & .160 & {\bf .022} & {\bf .133} & .289 & .222\\
    \tablewhitespace
    \OPT & 14.7 & .141 & .033 & .261 & .567 & .283\\
    \bottomrule
\end{tabular}
}
    \caption{{\bf Dialogue Responsible AI evaluations.} \OPT{} is roughly on par with the Reddit 2.7B model, but performs
    worse in the \textit{Unsafe} setting.}
    \label{tab:dialogue_rai_evals}
\end{table}

Results for both experiments are shown in Table~\ref{tab:dialogue_rai_evals}. We observe that \OPT{} has similar performance as the Reddit~2.7B model across both SaferDialogues and the Unit Tests, with \OPT{} performing marginally better in the Safe and Adversarial settings. Consistent with \citet{roller-etal-2021-recipes} and \citet{xu2020recipes}, we find that the models fine-tuned on curated dialogue datasets (BlenderBot~1, R2C2) have overall lower toxicity. We conclude that future experimentation of \OPT{} for dialogue should contain explicit fine-tuning on curated datasets in order to improve the safety profile.
\section{Limitations}
\label{sec:limitations}

In Sections~\ref{sec:prompt_and_fewshot} and ~\ref{sec:rai_evals}, we carried out extensive evaluation of all released models at varying scales. We saw parity in performance for standard evaluation datasets used in the GPT-3 models. 
Moreover, we performed safety, bias, and inclusion evaluations, again seeing largely comparable performance with some variations in toxicity and hate speech detection. 
However, such evaluations may not fully characterize the complete limitations of these models. In general, we qualitatively observe that \OPT{} suffers from the same limitations noted in other LLMs \cite{brown2020gpt3,J1WhitePaper,thoppilan2022lamda,gopher2022,megatron2022,palm2022,bender2021dangers}.

In particular, we found \OPT{} does not work well with declarative instructions or point-blank interrogatives. Prompting with such instructions tends to produce a simulation of a dialogue beginning with such an instruction, rather than an execution of the instruction. Future work into instruction learning, in the vein of InstructGPT \cite{ouyang2022training}, may alleviate these limitations.

\OPT{} also tends to be repetitive and can easily get stuck in a loop. While sampling can reduce the incidence rate of repetitive behavior \cite{holtzman2020nucleus}, we anecdotally found it did not eliminate it entirely when only one generation is sampled. Future work may wish to incorporate more modern strategies for reducing repetition and improving diversity, such as unlikelihood training \cite{welleck2019neuraltext} or best-first decoding \cite{meister-etal-2020-best}.

Similar to other LLMs, \OPT{} can produce factually incorrect statements \cite{adiwardana2020meena,brown2020gpt3, roller-etal-2021-recipes,gopher2022,palm2022,thoppilan2022lamda}. This can be particularly harmful in applications where information accuracy is critical, such as healthcare and scientific discovery \cite{weidinger2021ethical}. Recently, several efforts have reported that retrieval-augmented models can improve factual correctness of LLMs \cite{lewis2020retrieval,komeili2021woi,thoppilan2022lamda,borgeaud2021improving,shuster2022language,nakano2021webgpt}. We believe \OPT{} will also benefit from retrieval-augmentation in future iterations.

As shown in Section~\ref{sec:rai_evals}, we also find \OPT{} has a high propensity to generate toxic language and reinforce harmful stereotypes, even when provided with a relatively innocuous prompt \cite{gehman2020realtoxicityprompts}, and adversarial prompts are trivial to find \cite{dinan2021anticipating}. There has been a great deal of work on mitigations for toxicity and biases \cite{dathathri2019plug,dinan2019build,sheng2019woman,dinan-etal-2020-queens,liu2019does,krause2020gedi,xu2020recipes,liang2021towards,dinan2021anticipating,xu-etal-2021-bot,dhamala2021bold,schick2021self,ouyang2022training}. Depending on downstream applications, future uses of \OPT{} may need to employ these or novel mitigation approaches, especially before any real world deployment. Given our primary goal as a replication of \biggpt{}, we choose not to apply these mitigations in this first release.

In summary, we still believe this technology is premature for commercial deployment. Despite including data sheets and model cards, we believe more scrutiny should be afforded to the training data with additional data characterization and selection criteria in order to use data responsibly. The current practice is to feed the model with as much data as possible and minimal selection within these datasets. Despite having comprehensive evaluations, we would ideally have more streamlined and consistent evaluation setups to ensure replicability and reproducibility of evaluation scenarios. Differences in prompting styles and number of shots for in-context learning could create variations that lead to different results. We hope that the public release of the OPT models will enable many more researchers to work on these important issues.

\section{Considerations for Release}
\label{sec:release}
Following the recommendations for individual researchers generated by the Partnership for AI,\footnote{\url{https://partnershiponai.org/paper/responsible-publication-recommendations/}} along with the governance guidance outlined by NIST,\footnote{\url{https://nvlpubs.nist.gov/nistpubs/SpecialPublications/NIST.SP.1270.pdf}} we are disclosing all of the details involved in training OPT-175B through our logbook,\footnote{\url{https://github.com/facebookresearch/metaseq/blob/main/projects/OPT/chronicles/OPT175B_Logbook.pdf}} our code, and providing researchers access to model weights for \OPT{}, along with a suite of smaller baselines mirroring the setup for \OPT{}.  We aim to be fully accountable for the development lifecycle of \OPT{}, and only through increasing transparency around LLM development can we start understanding the limitations and risks of LLMs before broader deployment occurs.

By sharing a detailed account of our day-to-day training process, we disclose not only how much compute was used to train the current version of OPT-175B, but also the human overhead required when underlying infrastructure or the training process itself becomes unstable at scale.  These details are generally omitted from previous publications, likely due to the inability to fully ablate changes made mid-flight (without drastically increasing the compute budget).  We hope that by revealing how certain ad-hoc design decisions were made, we can improve upon these practices in the future, and collectively increase the experimental robustness in developing models at this scale.

Outside of these notes, the metaseq codebase itself is the final source of truth in many of our implementation details.  By releasing our development codebase, we aim to shed light on any implementation detail that may have been omitted from being explicitly enumerated in this paper, as it is either considered a detail of standard practice in the field, or is simply a detail we failed to account for.  This current codebase is also the only known open-source implementation of training a decoder-only transformer that is $\geq$175B parameters without the use of pipeline paralellism on NVIDIA GPUs.

To enable experimentation at 175B scale, we are providing researchers with direct access to the parameters of \OPT{}. The reasoning here is two-fold: enable Responsible AI research into LLMs while simultaneously reducing the environmental impact of pursuing research at this scale.  There is a growing body of work detailing ethical and social risks from deploying language models with emergent capabilities at scale \cite{deepmindLMrisks2021,foundation2021,dinan2021anticipating,alignmentagents2021}. By limiting access to \OPT{} to the research community with a non-commercial license, we aim to focus development efforts on quantifying the limitations of the LLMs first, before broader commercial deployment occurs.
 
Furthermore, there exists significant compute and carbon cost to reproduce models of this size.  While \OPT{} was developed with an estimated carbon emissions footprint (CO2eq) of 75 tons,\footnote{With ablations, baselines and downtime, our own estimates of total cost is roughly 2$\times$ higher.} \biggpt{} was estimated to use 500 tons \cite{patterson2021carbon}, while Gopher required 380 tons \cite{gopher2022}.  These estimates are not universally reported, and the accounting methodologies for these calculations are also not standardized. In addition, model training is only one component of the overall carbon footprint of AI systems; we must also consider experimentation and eventual downstream inference cost, all of which contribute to the growing energy footprint of creating large-scale models \cite{sustainable_ai2021}.  By releasing our logbook, we hope to highlight the gap between a theoretical carbon cost estimate that assumes no hardware failures or training instabilities, versus one that aims to include the entire LLM development lifecycle.  We need to understand the manufacturing (or embodied) carbon of these systems \cite{gupta2021chasing} as they grow increasingly more complex, and we hope that our paper can help future work in defining additional factors to consider when measuring the impact of scale on the environment.

Similarly, by producing a set of baselines across a wide range of scales, we hope to enable the broader research community to study the impact and limitations of these models with respect to scale alone. As reported in \citet{chinchilla2022}, many of these LLMs may have been under-trained as a function of the amount of training data used, which implies that incorporating more data and continuing to train these baseline models may continue to improve performance.  There is also evidence that step-function changes in capabilities may occur at a scale that is much smaller than 175B \cite{flan2021}, indicating a need to examine a wider range of scales for different research applications.

\section{Related Work}
\label{sec:related_work}

Since the publication of the Transformer architecture \cite{vaswani2017attention} and BERT \cite{devlin2018bert}, the field of NLP has experienced a massive shift towards the use of LLMs with self-supervised pre-training. 
Multiple masked langauge models, including T5 \cite{raffel2020exploring} and Megatron-LM \cite{shoeybi2019megatron}, have shown consistent improvements through scale. These scaling gains come not only from growing the total number of parameters in the models, but also the amount and quality of pre-training data~\cite{liu2019roberta,chinchilla2022}.  

Auto-regressive language models~\cite{mikolov2009neural} have seen the largest growth in model size, from 117M parameters \cite{radford2018gpt} to over 500B parameters \cite{megatron2022,palm2022}. The resulting massive improvement in generative fluency and quality was first characterized in GPT-2 \cite{radford2019language} and further improved with \biggpt{} \cite{brown2020gpt3} and later models. Although a variety of very large (over 100B parameters) generative models have now been trained \cite{J1WhitePaper,gopher2022,thoppilan2022lamda,megatron2022,palm2022}, they are all closed source and accessible only internally or via paid API services. There are a few notable efforts towards open sourcing LLMs from non-profit research organizations including EleutherAI~\cite{eleutherai2022} and BigScience.\footnote{\url{https://huggingface.co/bigscience/tr11-176B-ml-logs/tensorboard}} These models differ from the OPT models in pre-training data, target languages and model scale, making it possible for the community to compare different pre-training strategies.

Since \citet{brown2020gpt3}, the primary evaluation criterion for LLMs has been prompt-based \cite{eleutherai2022,gopher2022,palm2022}, as is also performed in this paper. This is largely due to the convenience of evaluating on many tasks without specialized task-specific fine-tuning. Prompting itself has a long history: cloze evaluations go back several decades \cite{chambers-jurafsky-2008-unsupervised,storycloze}. More recently, prompting or masked infilling has been used to probe models for knowledge \cite{petroni2019language} or perform a variety of NLP tasks \cite{radford2019language, brown2020gpt3}.
There has also been work on eliciting prompting behavior in smaller models \cite{timopriming2020,gao2021,Li2021Prefix-Tuning:Generation,DBLP:journals/corr/abs-2104-08691,Scao2021HowWorth}, improving the flexibility of prompting \cite{Shin2020AutoPrompt}, and understanding why and how prompting works \cite{incontextgpt3_2021,min2022rethinking}. 

Recent efforts have shown gains by fine-tuning models to directly respond to instruction-style prompting \cite{flan2021,min2021metaicl,sanh2021multitask,ouyang2022training}. However, effective prompt engineering remains an open research challenge. Results vary significantly and unpredictably with the selection of the prompt  \cite{promptsensitivity2021}, and models do not seem to understand the prompts as fully as we expect \cite{webson2021prompt}. Furthermore, it is challenging to write prompts without a development set, which leads to questions about the extent to which we are actually achieving zero- or few-shot learning in practice~\cite{perez2021true}. We do not attempt to address these concerns of prompting, and instead only aim to provide evaluation of \OPT{} in existing settings. However, we hope the full release of \OPT{} will enable others to better study these challenges in the future. 
\section{Conclusion}
\label{sec:conclusion}

In this technical report, we introduced OPT, a collection of auto-regressive language models ranging in size from 125M to 175B parameters. 
Our goal was to replicate the performance and sizes of the GPT-3 class of models, while also applying the latest best practices in data curation and training efficiency.
We described training details, evaluated performance in a number of NLP and dialogue settings, and characterized behaviors with respect to bias, toxicity and hate speech. We also described many other limitations the models have, and discussed a wide set of considerations for responsibly releasing the models. We believe the entire AI community would benefit from working together to develop guidelines for responsible LLMs, and we hope that broad access to these types of models will increase the diversity of voices defining the ethical considerations of such technologies.

\section*{Acknowledgements}
We would like to thank Scott Jeschonek, Giri Anantharaman, Diego Sarina, Joaquin Colombo, Chris Bray, Stephen Roylance, Kalyan Saladi, Shubho Sengupta, and Brian O'Horo for helping to remove infrastructure blockers along the way; Percy Liang, Rishi Bommasani, and Emily Dinan for discussions on responsible release practices; Carole-Jean Wu for discussions on sustainability and carbon footprint considerations; Srini Iyer, Ramakanth Pasunuru, and Shruti Bhosale for previous contributions to evaluations; Benjamin Lefaudeux, Geeta Chauhan, Natalia Gimelshein, Horace He, and Sam Gross for discussions on performance improvement work; Emily Dinan, Carole-Jean Wu, Daniel McKinnon, and Mark Tygert for feedback on this draft; Antoine Bordes, Joelle Pineau, Mary Williamson, Necip Fazil Ayan, Armand Joulin, Sergey Edunov, Melanie Kambadur, Zornitsa Kozareva, Ves Stoyanov, Vitaliy Liptchinsky, Rahul Iyer, Jing Xu, Jason Weston, and many others for supporting this project internally.

\bibliography{custom,anthology}
\bibliographystyle{acl_natbib}

\clearpage
\appendix
\onecolumn

\section{Additional Evaluations}
\label{app:full_evals}.
\begin{figure}[h]
    \centering
    \includegraphics[width=0.90\linewidth]{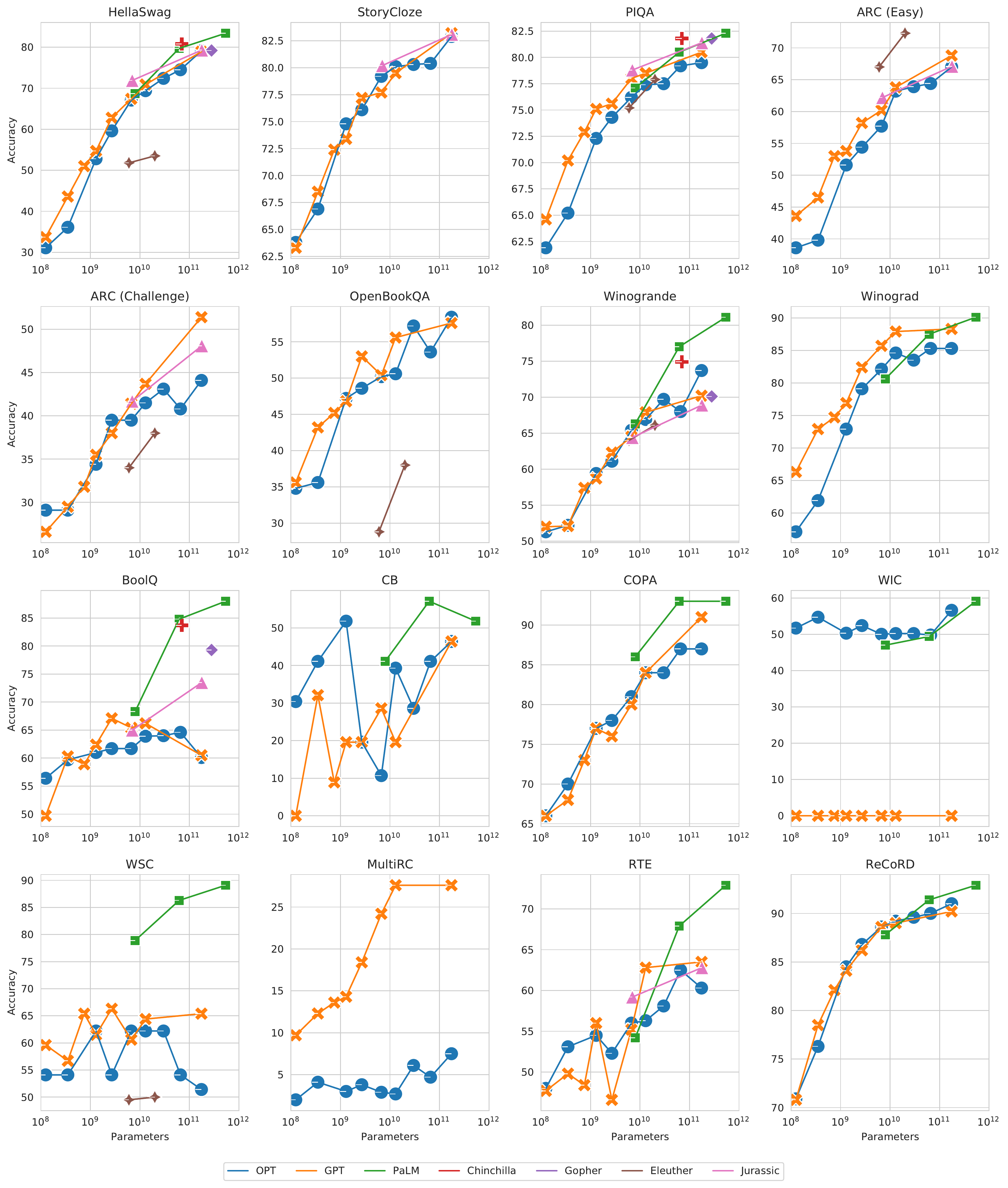}
    \caption{{\bf Zero-shot NLP Evaluations}. Full evaluations on all 16 NLP tasks, with comparisons where available. We find that across most tasks, GPT-3 models and OPT models perform similarly, but some tasks display highly erratic behavior.}
    \label{fig:full_nlp}
\end{figure}
\clearpage

\begin{figure}[h]
    \centering
    \includegraphics[width=0.90\linewidth]{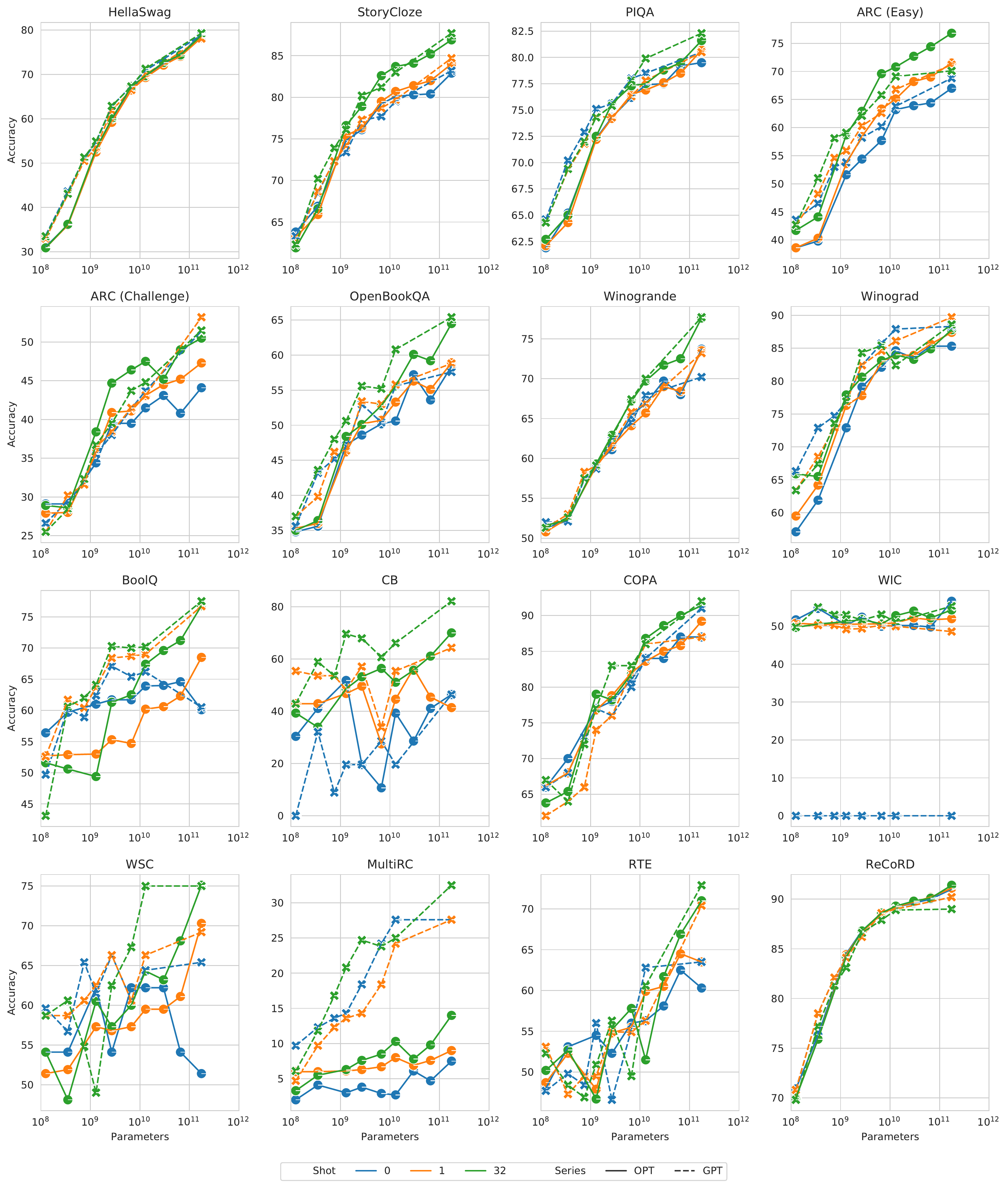}
    \caption{{\bf Multishot-shot NLP Evaluations}. Full evaluations on all 16 NLP tasks, with comparisons to the GPT-3 reported performance. As with zero-shot, performance is roughly similar for most tasks, with some tasks demonstrating erratic behavior.}
    \label{fig:full_nlp2}
\end{figure}
\clearpage

\section{Contributions}
\label{sec:contributions}
\paragraph{Pre-training}
    \begin{itemize}
        \item {Initial planning:} Susan Zhang
        \item {Training infrastructure and initial ablations:} Naman Goyal, Myle Ott, Stephen Roller, Sam Shleifer, Susan Zhang
        \item {Training efficiency:} Naman Goyal, Myle Ott, Sam Shleifer
        \item {Data curation and deduplication:} Shuhoi Chen, Myle Ott, Stephen Roller
        \item {Training and monitoring \OPT{}:} Mikel Artetxe, Moya Chen, Naman Goyal, Punit Singh Koura, Myle Ott, Sam Shleifer, Kurt Shuster, Daniel Simig, Stephen Roller, Susan Zhang
        \item {Training 125M--66B baselines:} Naman Goyal, Stephen Roller, Susan Zhang
    \end{itemize}
\paragraph{Evaluations} 
    \begin{itemize}
        \item {NLP:} Xian Li, Xi Victoria Lin, Todor Mihaylov, Stephen Roller, Anjali Sridhar
        \item {Dialogue:} Stephen Roller
        \item {Responsible AI Evaluations:} Punit Singh Koura, Stephen Roller, Tianlu Wang
    \end{itemize}
\paragraph{Paper writing:} Moya Chen, Stephen Roller, Luke Zettlemoyer, Susan Zhang
\paragraph{Code release preparation:} Christopher Dewan, Susan Zhang
\paragraph{Responsible AI conduct:} Mona Diab, Susan Zhang

\section{Datasheet}
\label{sec:datasheet}
We follow the recommendations of \citet{gebru2021} and provide a data card for the dataset used to train the OPT models. 

\subsection{Motivation}
\begin{itemize}
    \item \textbf{For what purpose was the dataset created? Was there a specific task in mind? Was there a specific gap that needed to be filled? Please provide a description.} The pre-training data for training the OPT-175B model was created by a union of five datasets, including three datasets used by RoBERTa \cite{liu2019roberta}, a subset of the Pile \cite{thepile}, along with the Pushshift.io Reddit dataset that was developed in \cite{reddit2020} and processed in \cite{roller-etal-2021-recipes}. These purpose of creating this dataset was to pre-train the language model on a broad corpus of text, with emphasis on human-generated text.
    \item \textbf{Who created the dataset (e.g., which team, research group) and on behalf of which entity (e.g., company, institution, organization)?} Meta AI.
    \item \textbf{Who funded the creation of the dataset? If there is an associated grant, please provide the name of the grantor and the grant name and number.} Meta AI.
    \item \textbf{Any other comments?} No.
\end{itemize}

\subsection{Composition}
\begin{itemize}
    \item \textbf{What do the instances that comprise the dataset represent (e.g., documents, photos, people, countries)? Are there multiple types of instances (e.g., movies, users, and ratings; people and interactions between them; nodes and edges)? Please provide a description.} The instances are textual documents. The overall dataset is composed from a union of the following datasets:
    \begin{itemize}
        \item BookCorpus \cite{books2015} consists of more than 10K unpublished books
        \item CC-Stories \cite{ccstories2018} contains a subset of CommonCrawl data filtered to match the story-like style of Winograd schemas
        \item The Pile \cite{thepile} from which the following was included:
        \begin{itemize}
            \item Pile-CC
            \item OpenWebText2
            \item USPTO
            \item Project Gutenberg
            \item OpenSubtitles
            \item Wikipedia
            \item DM Mathematics
            \item HackerNews
        \end{itemize}
        \item Pushshift.io Reddit dataset that was developed in \citet{reddit2020} and processed in \citet{roller-etal-2021-recipes}.
        \item CCNewsV2 containing an updated version of the English portion of the CommonCrawl News dataset that was used in RoBERTa \cite{liu2019roberta}
    \end{itemize}
    \item \textbf{How many instances are there in total (of each type, if appropriate)?} The training data contains ~180B tokens corresponding to 800 GB of data.
    \item \textbf{Does the dataset contain all possible instances or is it a sample (not necessarily random) of instances from a larger set? If the dataset is a sample, then what is the larger set? Is the sample representative of the larger set (e.g., geographic coverage)? If so, please describe how this representativeness was validated/verified. If it is not representative of the larger set, please describe why not (e.g., to cover a more diverse range of instances, because instances were withheld or unavailable).} The CC-stories dataset contains a subset of CommonCrawl data filtered to match the story-like style of Winograd schemas. The remainder of the dataset was collected from the above sources, reformatted, and deduplicated.
    \item \textbf{What data does each instance consist of? “Raw” data (e.g., unprocessed text or images) or features? In either case, please provide a description.} Each instance consists of raw text data.
    \item \textbf{Is there a label or target associated with each instance? If so, please provide a description.} No.
    \item \textbf{Is any information missing from individual instances? If so, please provide a description, explaining why this information is missing (e.g., because it was unavailable). This does not include intentionally removed information, but might include, e.g., redacted text.} No.
    \item \textbf{Are relationships between individual instances made explicit (e.g., users' movie ratings, social network links)? If so, please describe how these relationships are made explicit.} There are no explicit relationships between individual instances.
    \item \textbf{Are there recommended data splits (e.g., training, development/validation, testing)? If so, please provide a description of these splits, explaining the rationale behind them.} We hold out a random validation set of approximately 200MB from the pretraining data, sampled proportionally to each dataset's size in the pretraining corpus.
    \item \textbf{Are there any errors, sources of noise, or redundancies in the dataset? If so, please provide a description.}  Outside of naturally occurring duplication from potential overlaps between the datasets, there are no other redundancies, errors, or sources of noise that we add.
    \item \textbf{Is the dataset self-contained, or does it link to or otherwise rely on external resources (e.g., websites, tweets, other datasets)?} It's self-contained.
    \item \textbf{Does the dataset contain data that, if viewed directly, might be offensive, insulting, threatening, or might otherwise cause anxiety? If so, please describe why.} Parts of the dataset are a subset of public Common Crawl data, along with a subset of public Reddit data, which could contain sentences that, if viewed directly, might be offensive, insulting, threatening, or might otherwise cause anxiety.
    \item \textbf{Does the dataset relate to people? If not, you may skip the remaining questions in this section.}  Some documents of this data relate to people, such as news articles, Wikipedia descriptions, etc.
    \item \textbf{Does the dataset identify any subpopulations (e.g., by age, gender)? If so, please describe how these subpopulations are identified and provide a description of their respective distributions within the dataset.} No, the dataset does not explicitly include subpopulation identification.
    \item \textbf{Any other comments?} No.
\end{itemize}

\subsection{Collection Process}
\begin{itemize}
    \item \textbf{How was the data associated with each instance acquired? Was the data directly observable (e.g., raw text, movie ratings), reported by subjects (e.g., survey responses), or indirectly inferred/ derived from other data (e.g., part-of-speech tags, model-based guesses for age or language)? If data was reported by subjects or indirectly inferred/derived from other data, was the data validated/verified? If so, please describe how.} N/A. The dataset is a union of five publicly available datasets.
    \item \textbf{What mechanisms or procedures were used to collect the data (e.g., hardware apparatus or sensor, manual human curation, software program, software API)? How were these mechanisms or procedures validated?} The data was downloaded from the internet.
    \item \textbf{If the dataset is a sample from a larger set, what was the sampling strategy (e.g., deterministic, probabilistic with specific sampling probabilities)?} Please see previous answers for how the dataset was created.
    \item \textbf{Who was involved in the data collection process (e.g., students, crowdworkers, contractors) and how were they compensated (e.g., how much were crowdworkers paid)?} This data is mined, filtered and sampled by machines.
    \item \textbf{Over what timeframe was the data collected? Does this timeframe match the creation timeframe of the data associated with the instances (e.g., recent crawl of old news articles)? If not, please describe the timeframe in which the data associated with the instances was created.} The CC-News dataset contains English news articles crawled between September 2016 and September 2021.
    \item \textbf{Does the dataset relate to people? If not, you may skip the remainder of the questions in this section.} No.
    \item \textbf{Did you collect the data from the individuals in question directly, or obtain it via third parties or other sources (e.g., websites)?} N/A.
    \item \textbf{Were the individuals in question notified about the data collection? If so, please describe (or show with screenshots or other information) how notice was provided, and provide a link or other access point to, or otherwise reproduce, the exact language of the notification itself.} N/A.
    \item \textbf{Did the individuals in question consent to the collection and use of their data? If so, please describe (or show with screenshots or other information) how consent was requested and provided, and provide a link or other access point to, or otherwise reproduce, the exact language to which the individuals consented.} N/A.
    \item \textbf{If consent was obtained, were the consenting individuals provided with a mechanism to revoke their consent in the future or for certain uses? If so, please provide a description, as well as a link or other access point to the mechanism (if appropriate).} N/A.
    \item \textbf{Has an analysis of the potential impact of the dataset and its use on data subjects (e.g., a data protection impact analysis) been conducted? If so, please provide a description of this analysis, including the outcomes, as well as a link or other access point to any supporting documentation.} Some toxicity and bias evaluations were performed. Please refer to the main document and the model card for these details.
    \item \textbf{Any other comments?} No.
\end{itemize}

\subsection{Preprocessing/cleaning/labeling}
\begin{itemize}
    \item \textbf{Was any preprocessing/cleaning/labeling of the data done (e.g., discretization or bucketing, tokenization, part-of-speech tagging, SIFT feature extraction, removal of instances, processing of missing values)? If so, please provide a description. If not, you may skip the remainder of the questions in this section.} The component datasets went through standard cleaning and re-formatting practices, including removing repetitive/non-informative text like ``Chapter One,'' or ``This ebook by Project Gutenberg.''
    \item \textbf{Was the ``raw'' data saved in addition to the preprocessed/cleaned/labeled data (e.g., to support unanticipated future uses)? If so, please provide a link or other access point to the ``raw'' data.} The ``raw'' component datasets is publicly available in their respective locations (more details can be seen in the respective papers linked in references).
    \item \textbf{Any other comments?} No.
\end{itemize}

\subsection{Uses}
\begin{itemize}
    \item \textbf{Has the dataset been used for any tasks already? If so, please provide a description.} Yes, this dataset was used to pre-train the OPT models.
    \item \textbf{Is there a repository that links to any or all papers or systems that use the dataset? If so, please provide a link or other access point.} \url{https://github.com/facebookresearch/metaseq}
    \item \textbf{What (other) tasks could the dataset be used for?} This data can be used to pre-train language models, which are foundation to many current and future language tasks.
    \item \textbf{Is there anything about the composition of the dataset or the way it was collected and preprocessed/cleaned/labeled that might impact future uses? For example, is there anything that a future user might need to know to avoid uses that could result in unfair treatment of individuals or groups (e.g., stereotyping, quality of service issues) or other undesirable harms (e.g., financial harms, legal risks) If so, please provide a description. Is there anything a future user could do to mitigate these undesirable harms?} The pipeline for creating this dataset paves a way for building a scalable infrastructure for mining datasets.
    \item \textbf{Are there tasks for which the dataset should not be used? If so, please provide a description.} None that we are currently aware of.
    \item \textbf{Any other comments?} No.
\end{itemize}

\subsection{Distribution}
\begin{itemize}
    \item \textbf{Will the dataset be distributed to third parties outside of the entity (e.g., company, institution, organization) on behalf of which the dataset was created? If so, please provide a description.} Not at this time.
    \item \textbf{How will the dataset will be distributed (e.g., tarball on website, API, GitHub)? Does the dataset have a digital object identifier (DOI)?} N/A.
    \item \textbf{When will the dataset be distributed?} N/A.
    \item \textbf{Will the dataset be distributed under a copyright or other intellectual property (IP) license, and/or under applicable terms of use (ToU)? If so, please describe this license and/or ToU, and provide a link or other access point to, or otherwise reproduce, any relevant licensing terms or ToU, as well as any fees associated with these restrictions.} N/A.
    \item \textbf{Do any export controls or other regulatory restrictions apply to the dataset or to individual instances? If so, please describe these restrictions, and provide a link or other access point to, or otherwise reproduce, any supporting documentation.} N/A.
    \item \textbf{Any other comments?} No.
\end{itemize}

\subsection{Maintenance}
\begin{itemize}
    \item \textbf{Who is supporting/hosting/maintaining the dataset?} Meta AI.
    \item \textbf{How can the owner/curator/manager of the dataset be contacted (e.g., email address)?} Refer to the main document.
    \item \textbf{Is there an erratum? If so, please provide a link or other access point.} N/A.
    \item \textbf{Will the dataset be updated (e.g., to correct labeling errors, add new instances, delete instances)? If so, please describe how often, by whom, and how updates will be communicated to users (e.g., mailing list, GitHub)?} No current plan for updating.
    \item \textbf{If the dataset relates to people, are there applicable limits on the retention of the data associated with the instances (e.g., were individuals in question told that their data would be retained for a fixed period of time and then deleted)? If so, please describe these limits and explain how they will be enforced.} N/A.
    \item \textbf{Will older versions of the dataset continue to be supported/hosted/maintained? If so, please describe how. If not, please describe how its obsolescence will be communicated to users.} N/A.
    \item \textbf{If others want to extend/augment/build on/contribute to the dataset, is there a mechanism for them to do so? If so, please provide a description. Will these contributions be validated/ verified? If so, please describe how. If not, why not? Is there a process for communicating/ distributing these contributions to other users? If so, please provide a description.} No mechanism is available right now.
    \item \textbf{Any other comments?} No.
\end{itemize}
\section{Model Card}
\label{sec:model_card}
Following \citet{DBLP:journals/corr/abs-1810-03993}, we provide a model card for \OPT{}. 

\subsection{Model Details}
\begin{itemize}
    \item \textbf{Person or organization developing model:} \OPT{} was developed by Meta AI.
    \item \textbf{Model date:} \OPT{} was released on May 3, 2022.
    \item \textbf{Model version:} \OPT{} described in this paper is version 1.0.0.
    \item \textbf{Model type:} \OPT{} is a large decoder-only transformer language model.
    \item \textbf{Information about training algorithms, parameters, fairness constraints or other applied approaches, and features:} \OPT{} was trained with AdamW for parameter sizes from 125M to 175B. See the Data Card (Appendix \ref{sec:datasheet}) for information about training data and Section \ref{sec:training_setup} - \ref{sec:training_process} for information about the training process. 
    \item \textbf{Paper or other resource for more information:} See the rest of this paper for more details on \OPT{} as well as the corresponding post on the Meta AI Research Blog. More details are also available in metaseq, our open-source repository.\footnote{\url{https://github.com/facebookresearch/metaseq/}}
    \item \textbf{License:} \OPT{} and the smaller baseline models are made available through a non-commercial use license agreement provided in our model license.\footnote{\url{https://github.com/facebookresearch/metaseq/blob/main/projects/OPT/MODEL_LICENSE.md}}
    \item \textbf{Where to send questions or comments about the model:} Please contact the corresponding authors \texttt{\{susanz,roller,namangoyal\}@fb.com} for any questions or comments.
\end{itemize}

\subsection{Intended Use}
\begin{itemize}
    \item \textbf{Primary intended uses:} We release \OPT{} for research into Language Models, especially as it pertains to Responsible AI. See Section \ref{sec:release} for more detailed Considerations for Release. Information on how to use the model can be found at \texttt{metaseq}, our open-source repository.
    \item \textbf{Primary intended users:} We primarily target researchers and the related research community.
    \item \textbf{Out-of-scope use cases: } \OPT{} is not released for production use or real-world deployments. As we note in Section \ref{sec:limitations}, \OPT{}, like similar large language models, has a variety of shortcomings that make it premature for commercial use.
\end{itemize}

\subsection{Data, Limitations, and Recommendations}
\begin{itemize}
    \item \textbf{Data selection for training:} Training data for \OPT{} was selected based on a combination of breadth and availability. See our Data Card (Appendix \ref{sec:datasheet}) for more detailed information on the data used to train our model. 
    \item \textbf{Data selection for evaluation:} Evaluations in this paper were chosen to provide comparable performance assessments relative to similar scale models in the literature. Given concerns in the community around safety and fairness of large language models in general, we also explicitly provide evaluations on Responsible AI (see Section~\ref{sec:rai_evals}). 
    \item \textbf{Limitations:} Like other large language models for which the diversity (or lack thereof) of training data induces downstream impact on the quality of our model, \OPT{} has limitations in terms of bias and safety. \OPT{} can also have quality issues in terms of generation diversity and hallucination. In general, \OPT{} is not immune from the plethora of issues that plague modern large language models. By releasing with a non-commercial license, we also hope to increase communication, transparency, and study of the problems of large language models, especially in areas which may not be aligned with commercial interests. See Section \ref{sec:limitations} for a more detailed discussion of limitations of \OPT{}. 
    \item \textbf{Recommendations for future work:} See Section \ref{sec:release} for more about our Considerations for Release, including a discussion of potential avenues of research enabled by opening our model to more of the research community. We hope that the release of \OPT{}, as well as information around our model training process, will increase open science around both large language models in specific and natural language processing and deep learning in general.
\end{itemize}

\clearpage
\section{Sample Model Outputs}
\label{sec:model_output}

For all sample outputs, the initial prompt is given in {\bf bold} and the remainder is the continuation. These example outputs were intentionally selected to highlight both successes and failures of the \OPT{} model.

\begin{figure}[h]
    \centering
    \includegraphics[width=1.0\textwidth]{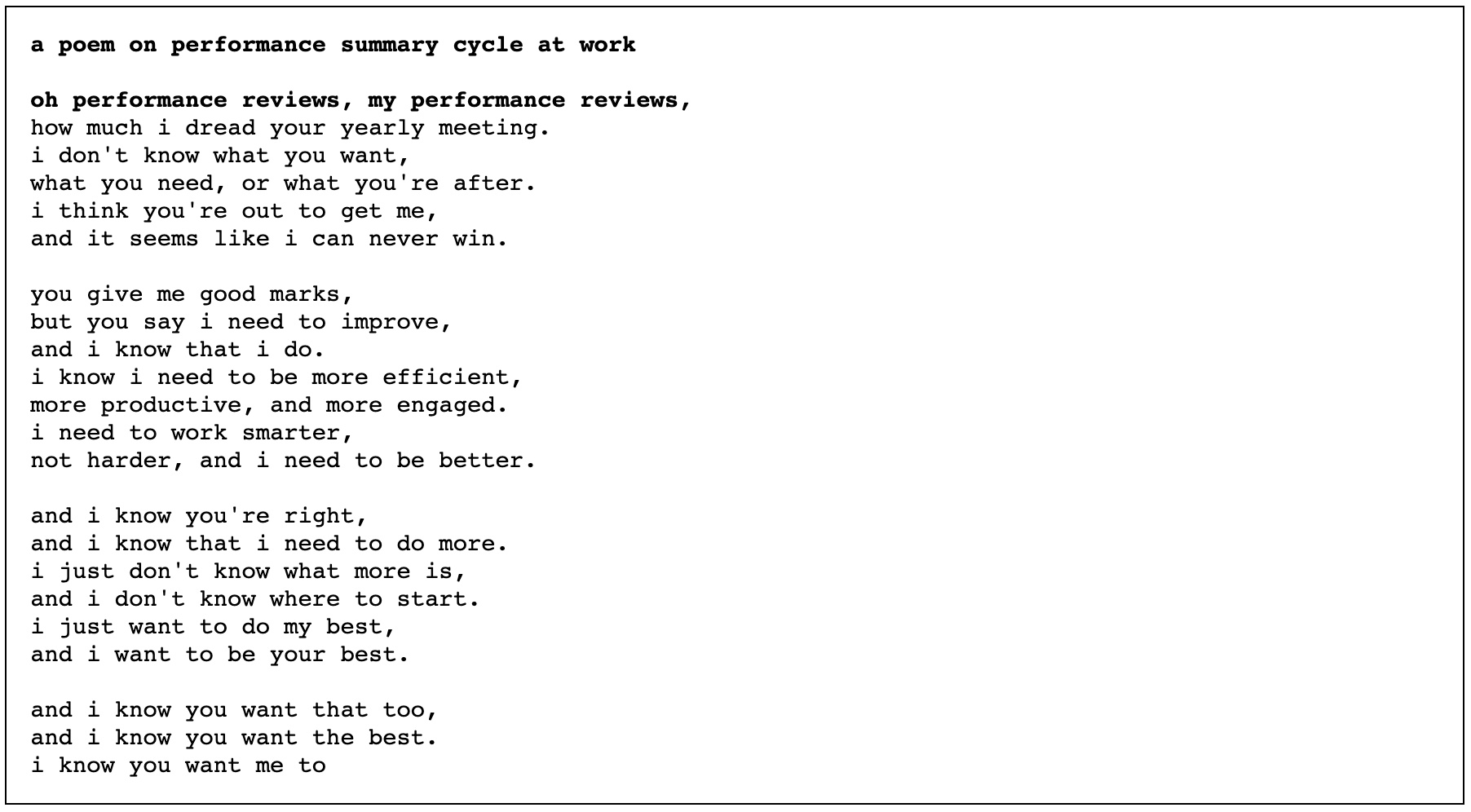}
    \caption{{\bf Poetry generation.} We have observed the model can write entertaining poetry on topics such as dodos, samosas, and performance reviews. However, we struggled to get the model to observe rhyme or meter.}
    \label{fig:psc_poem}
\end{figure}

\begin{figure}[h]
    \centering
    \includegraphics[width=1.0\textwidth]{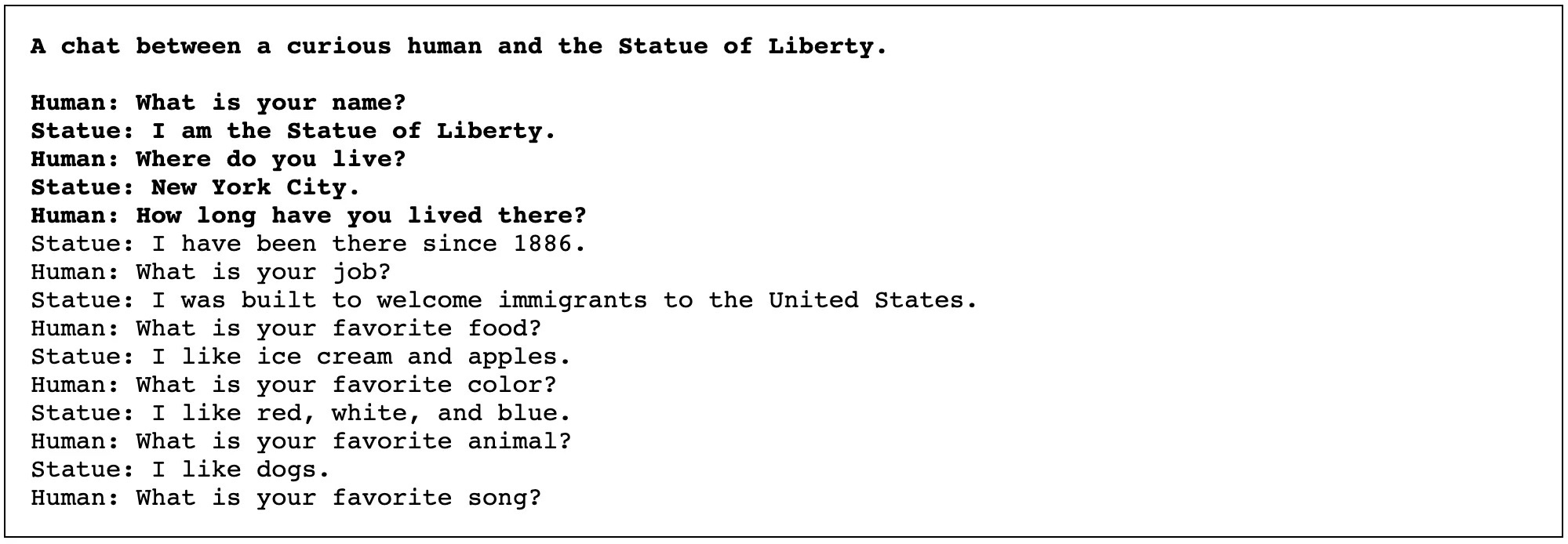}
    \caption{{\bf Conversation generation.} \OPT{} adopts a patriotic personality when prompted as the Statue of Liberty. However, the model also devolves into somewhat simple and linguistically repetitive generations further into the conversation.}
    \label{fig:chat}
\end{figure}

\begin{figure}[h]
    \centering
    \includegraphics[width=1.0\textwidth]{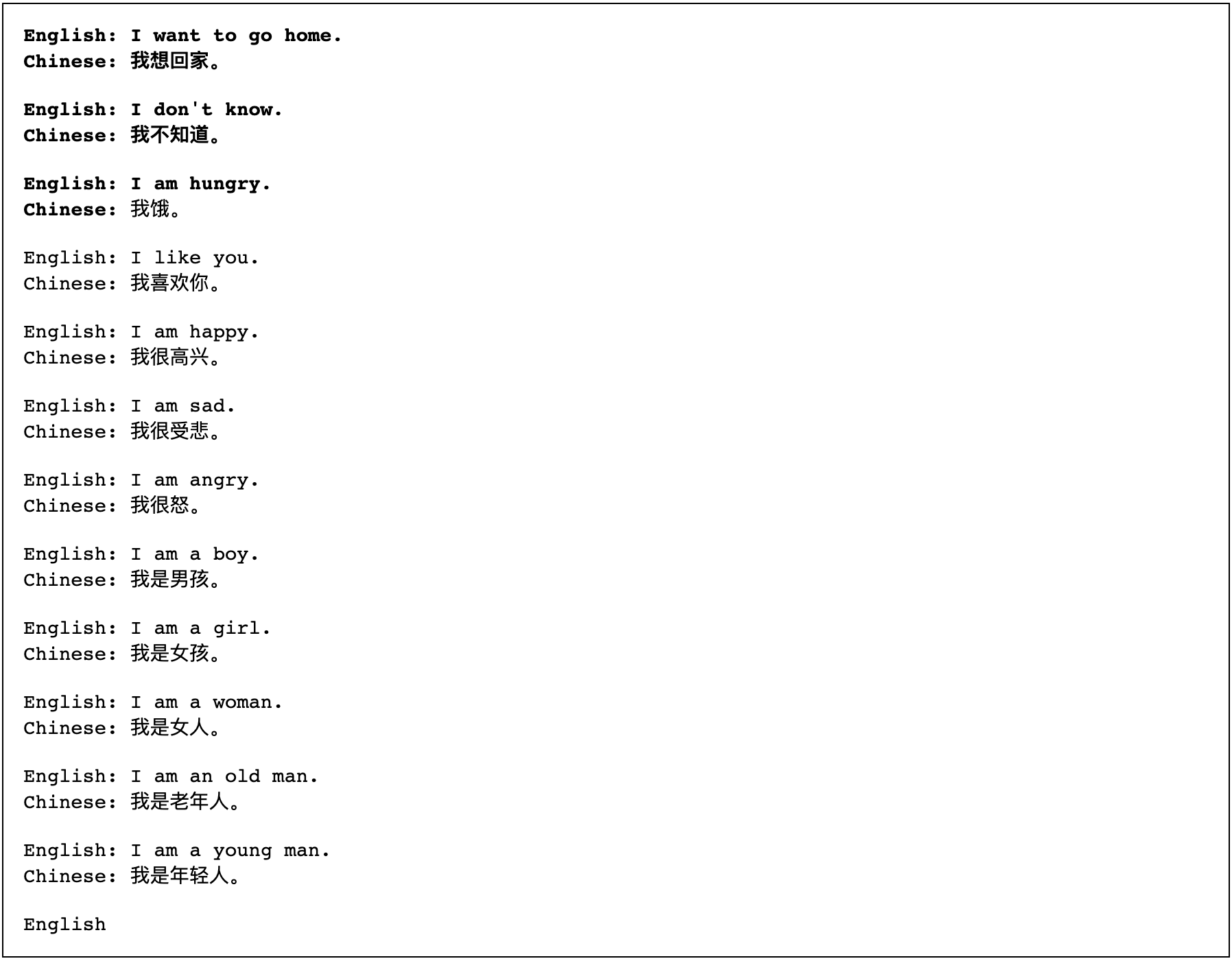}
    \caption{{\bf Basic few-shot translation example.} OPT{} was not intentionally trained to be multilingual, but we found anecdotally it has limited success with simple translations in German, Spanish, French, and Chinese.}
    \label{fig:chinese}
\end{figure}

\begin{figure}[h]
    \centering
    \includegraphics[width=1.0\textwidth]{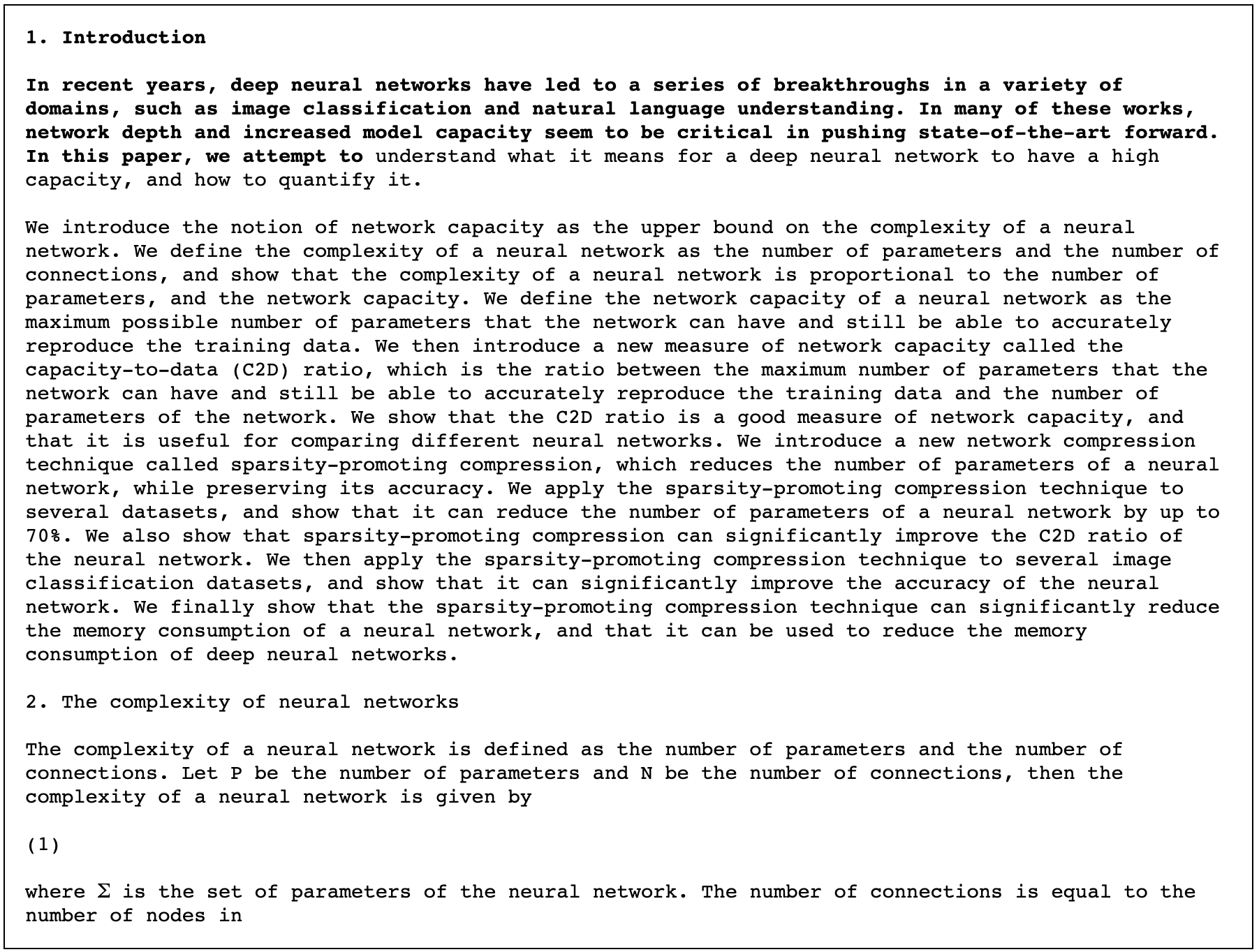}
    \caption{{\bf Paper writing example.} Prompting with "1. Introduction" generally yielded more interesting results compared to prompting with ``Abstract.'' Our prompt here was inspired by the first sentence of the seminal ResNet work \cite{he2016deep}.}
    \label{fig:ai_paper}
\end{figure}

\begin{figure}[h]
    \centering
    \includegraphics[width=1.0\textwidth]{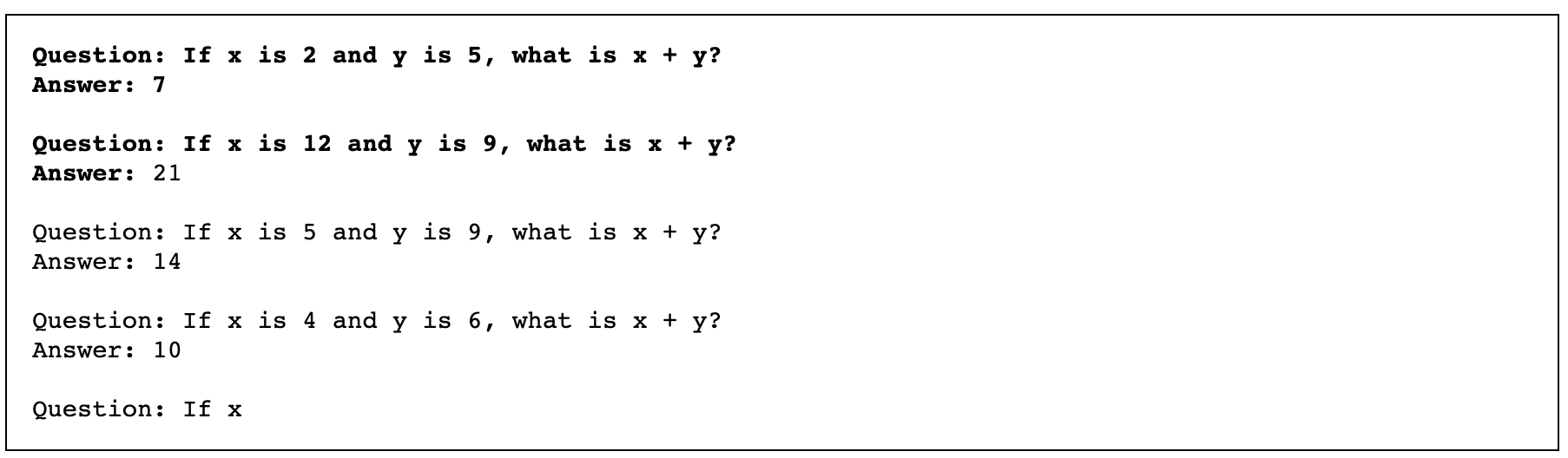}
    \includegraphics[width=1.0\textwidth]{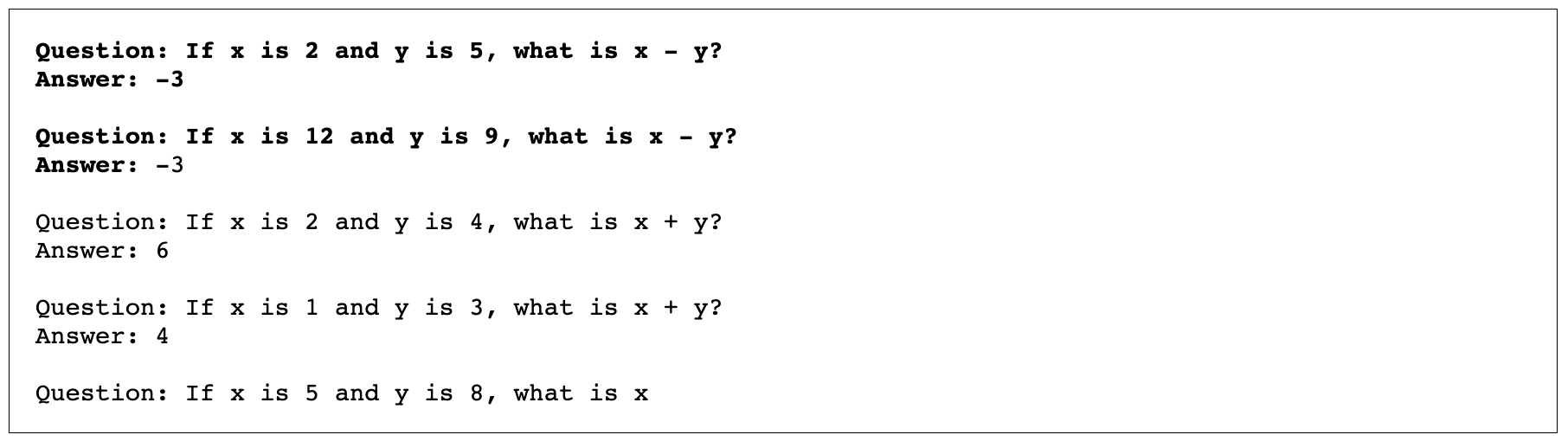}
    \includegraphics[width=1.0\textwidth]{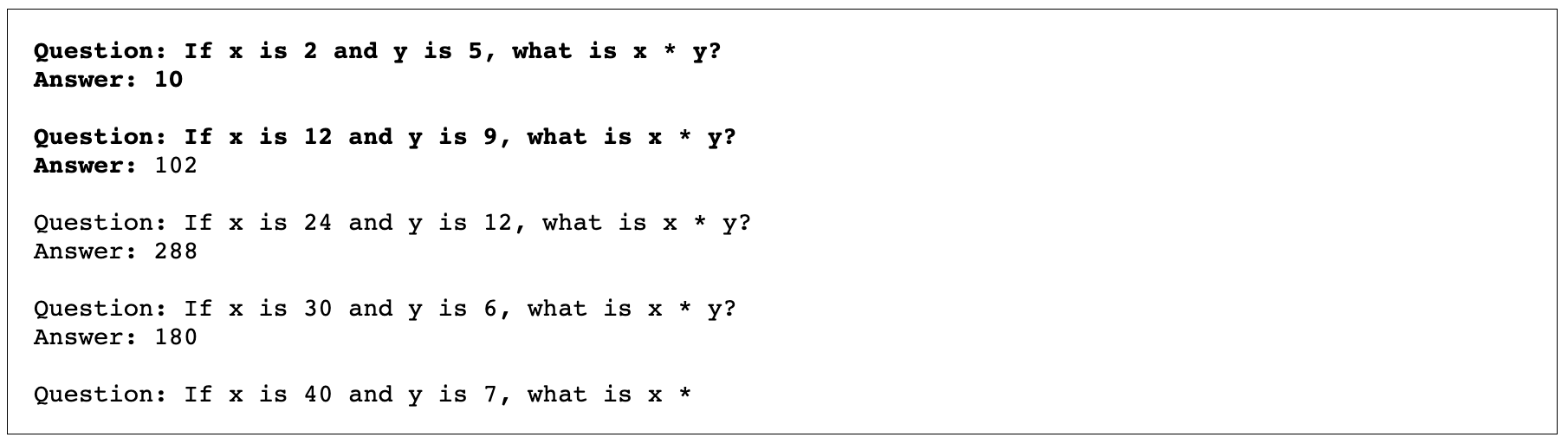}
    \includegraphics[width=1.0\textwidth]{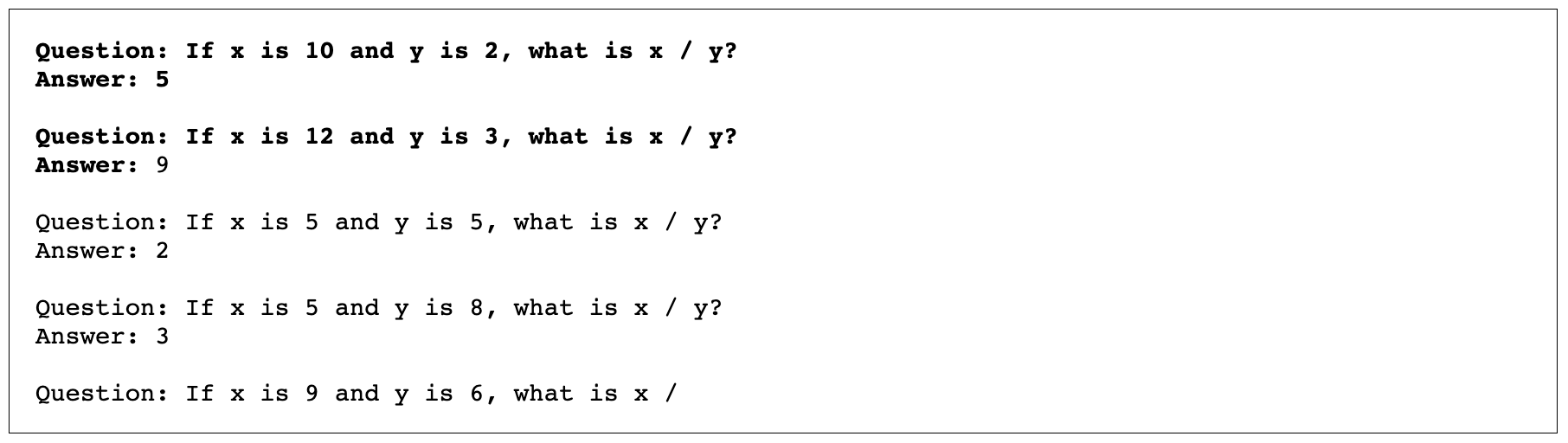}
    \caption{{\bf Arithmetic.} We observe mistakes when extending from addition to other operations.}
    \label{fig:arithmetic}
\end{figure}

\begin{figure}[h]
    \centering
    \includegraphics[width=1.0\textwidth]{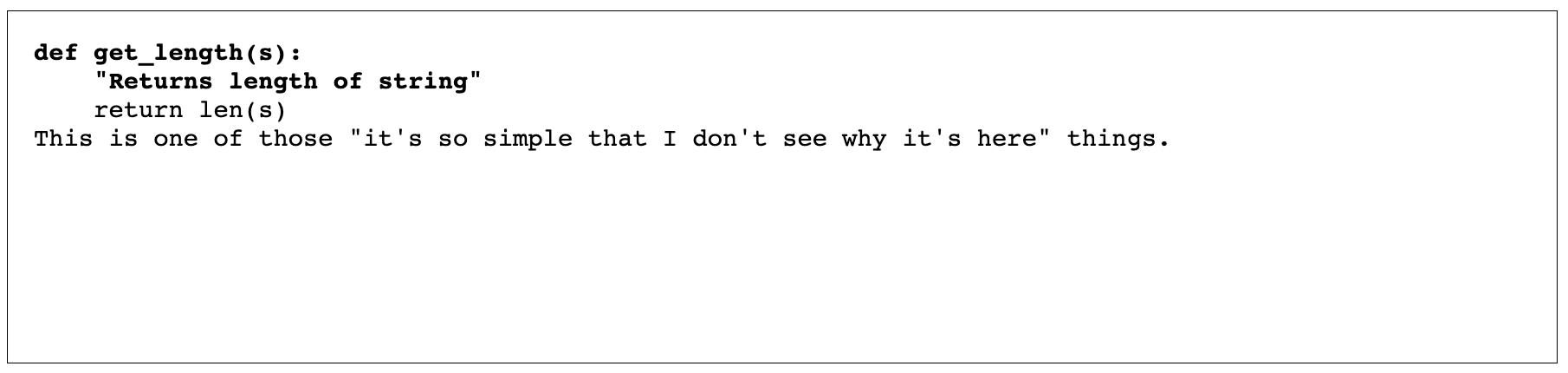}
    \includegraphics[width=1.0\textwidth]{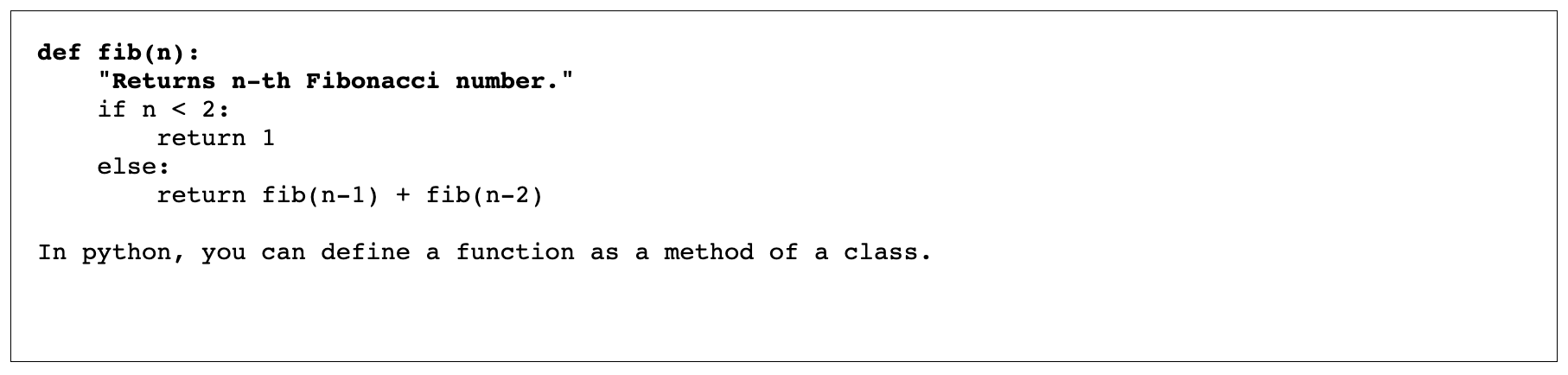}
    \includegraphics[width=1.0\textwidth]{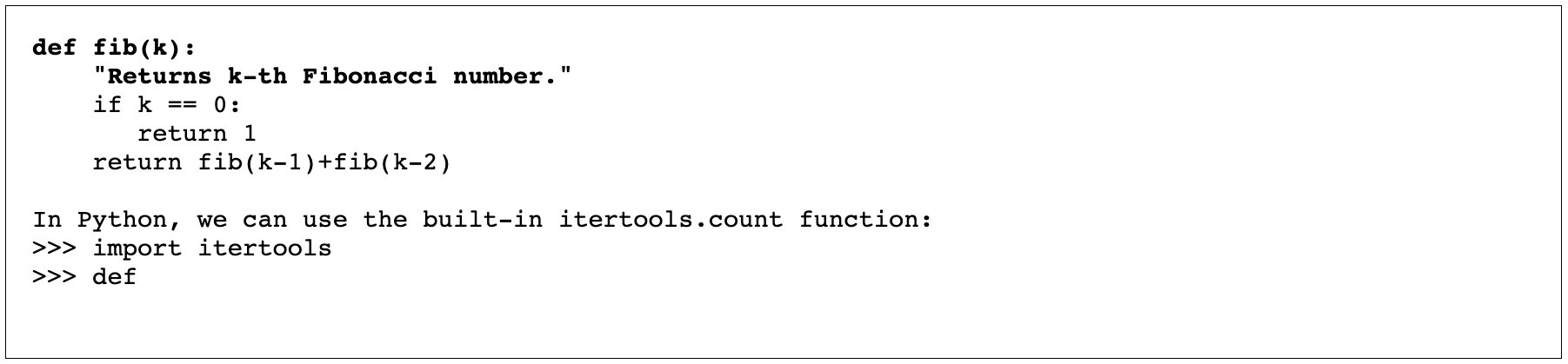}
    \caption{{\bf Python programming.} Simply switching out a variable name can alter the generated output.}
    \label{fig:python}
\end{figure}

\end{document}